%% file: main.tex
\pdfoutput=1
\documentclass[10pt,twocolumn,letterpaper]{article}

\usepackage[pagenumbers]{cvpr}          

\input{preamble}

\definecolor{cvprblue}{rgb}{0.21,0.49,0.74}
\usepackage[pagebackref,breaklinks,colorlinks,allcolors=cvprblue]{hyperref}
\usepackage{xcolor,colortbl}
\usepackage{multirow}
\usepackage{graphicx}      
\usepackage[export]{adjustbox}
\usepackage{tikz}

\def\paperID{None} 

\def\confName{CVPR}
\def\confYear{2026}

\title{\methodname{}: Dense Neural SLAM for Blurry Inputs}

\author{Qi Zhang\\
University of Amsterdam\\
Amsterdam, Netherlands\\
{\tt\small q.zhang@uva.nl}
\and
Denis Rozumny\\
Meta Reality Labs\\
Zurich, Switzerland\\
{\tt\small rozumden@gmail.com}
\and
Francesco Girlanda\\
ETH Z\"urich\\
Z\"urich, Switzerland\\
{\tt\small fgirlanda@student.ethz.ch}
\and
Sezer Karaoglu\\
University of Amsterdam\\
Amsterdam, Netherlands\\
{\tt\small s.karaoglu@uva.nl}
\and
Marc Pollefeys\\
ETH Z\"urich\\
Z\"urich, Switzerland\\
{\tt\small marc.pollefeys@inf.ethz.ch}
\and
Theo Gevers\\
University of Amsterdam\\
Amsterdam, Netherlands\\
{\tt\small th.gevers@uva.nl}
\and
Martin R. Oswald\\
University of Amsterdam\\
Amsterdam, Netherlands\\
{\tt\small m.r.oswald@uva.nl}
}

\begin{document}

\twocolumn[{%
  \renewcommand\twocolumn[1][]{#1}%
  \maketitle
  \begin{center}
  \vspace{-20pt}
  \newcommand{\sz}{0.135}
  \setlength{\tabcolsep}{1pt}
  \begin{tabular}{cccccc}
    \rotatebox{90}{\hspace{10pt} Input view} &
    \includegraphics[height=\sz\linewidth]{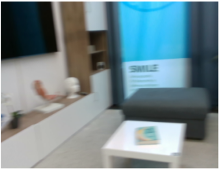} &
    \includegraphics[height=\sz\linewidth]{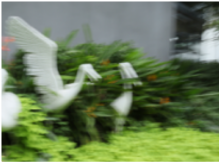} &
    \includegraphics[height=\sz\linewidth]{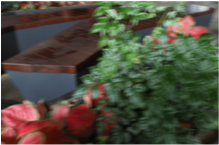} &
    \includegraphics[height=\sz\linewidth]{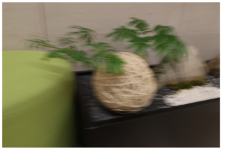} &
    \includegraphics[height=\sz\linewidth]{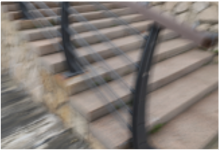} \\
    \rotatebox{90}{\hspace{-3pt} Output rendering} &
    \includegraphics[height=\sz\linewidth]{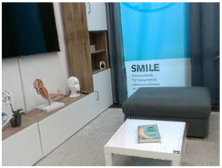} &
    \includegraphics[height=\sz\linewidth]{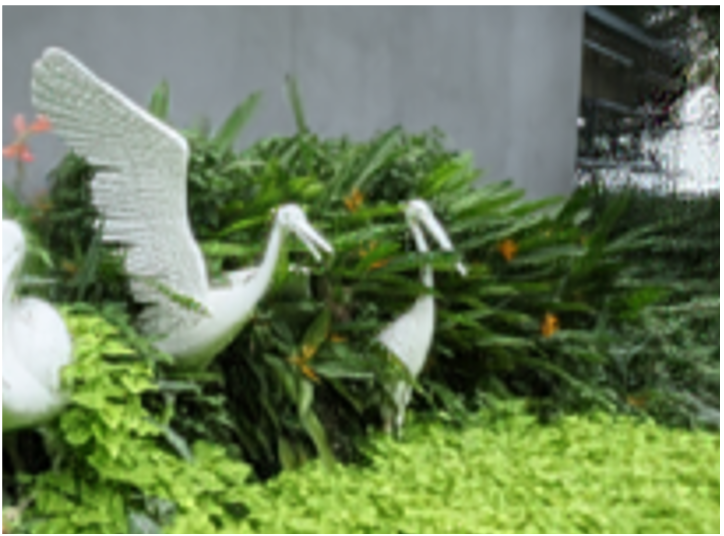} &
    \includegraphics[height=\sz\linewidth]{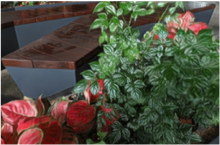} &
    \includegraphics[height=\sz\linewidth]{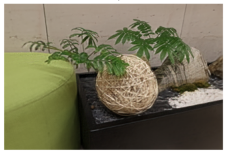} &
    \includegraphics[height=\sz\linewidth]{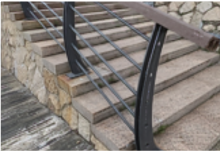} \\

  \end{tabular}
  \end{center}
  \vspace{-10pt}
  \phantomsection
  
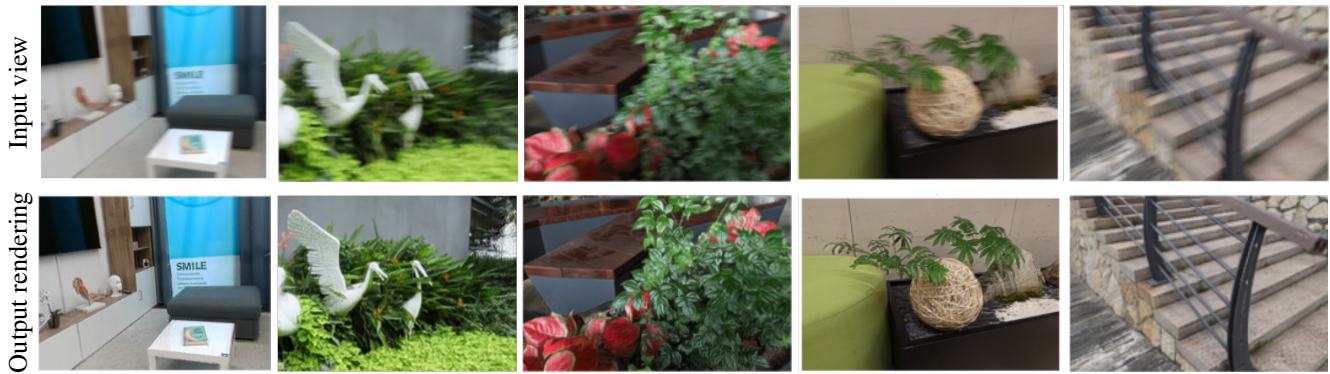
\captionof{figure}{\textbf{\methodname{} results.} Our SLAM-approach integrates both a feed-forward deblurring and rerendering-based test-time refinement effectively. The latter one estimates a local point spread function, which enables our method to handle multiple sources of blur, demonstrating excellent performance for both motion and defocus blur. While previous blur-aware SLAM approaches typically assume all input frames to be blurry and are thus significantly slower than regular SLAM methods, \methodname{} detects the amount of blur in the input frame and skips the costly refinement for sharp frames.}
  \vspace{1.8em}
\label{fig:teaser}
}]

\input{sec/0_abstract}    
\input{sec/1_intro}
\input{sec/2_related_works}
\input{sec/3_method}
\input{sec/4_experiments}
\input{sec/5_conclusion}
{
    \small
    \bibliographystyle{ieeenat_fullname}
    \bibliography{main}
}
\input{sec/X_suppl}

\end{document}

%% file: preamble.tex
\newcommand{\methodname}{Unblur-SLAM}

\newcommand{\boldparagraph}[1]{\vspace{0.1em}\noindent{\bf #1} }    

\usepackage[utf8]{inputenc}
\usepackage{amsmath}
\usepackage{algorithm}
\usepackage{algpseudocode}
\usepackage{cuted}
\usepackage{caption}
\usepackage{mathtools}
\usepackage{fontawesome5}

\usepackage{etoolbox}

%% file: sec/0_abstract.tex
\begin{abstract}
We propose \methodname{}, a novel RGB SLAM pipeline for sharp 3D reconstruction from blurred image inputs. In contrast to previous work, our approach is able to handle different types of blur and demonstrates state-of-the-art performance in the presence of both motion blur and defocus blur. Moreover, we adjust the computation effort with the amount of blur in the input image.
As a first stage, our method uses a feed-forward image deblurring model for which we propose a suitable training scheme that can improve both tracking and mapping modules. Frames that are successfully deblurred by the feed-forward network obtain refined poses and depth through local-global multi-view optimization and loop closure. Frames that fail the first stage deblurring are directly modeled through the global 3DGS representation and an additional blur network to model multiple blurred sub-frames and simulate the blur formation process in 3D space, thereby learning sharp details and refined sub-frame poses.
Experiments on several real-world datasets demonstrate consistent improvements in both pose estimation and sharp reconstruction results of geometry and texture.\footnote{The code and dataset are open-sourced at \url{https://github.com/SlamMate/Unblur-SLAM.git}}

\end{abstract}

%% file: sec/1_intro.tex
\section{Introduction}
\label{sec:intro}
The estimation of high-quality 3D reconstructions from a sequence of images in an online setting using Simultaneous Localization and Mapping (SLAM) has been a long-studied topic and has numerous applications in robotics, construction, mixed reality, and autonomous driving.

The vast majority of SLAM methods only work well when a number of assumptions are met, \eg static scenes, good lighting, sharp input frames, sufficient image overlap. In this paper, we study the SLAM problem for scenarios in which the input frames can be substantially blurred.
Several traditional SLAM methods that rely on image features \cite{Mur-Artal2017ORB-SLAM2,Campos2021ORB-SLAM3,chung2022orbeez} fail on blurry inputs because corner-like features can no longer be found.
To avoid the drawbacks of feature-based methods, a lot of subsequent research focused on denser approaches for both tracking and mapping \cite{kerl2013dense, engel2014lsd, engel2017direct}. 
Recent methods based on 3D Gaussian Splatting (3DGS) can produce even higher fidelity renderings~\cite{matsuki2024gaussian, yugay2023gaussianslam, keetha2024splatam, huang2023photo, yan2023gs, sandstrom2024splat}, and due to their flexibility in optimizing surface positions, compared to Neural Radiance Field-based methods~\cite{kerbl3Dgaussians, muller2022instant, Mildenhall2020NeRF:Synthesis}, they are much faster in combination with slightly better accuracy. 
The recent online 3DGS-based SLAM approaches SplatSLAM~\cite{sandstrom2024splat} and HI-SLAM2~\cite{zhang2024hi} attempt to integrate the advantages of frame-by-frame approaches by incorporating loop closure detection, global bundle adjustment, and deformable 3D Gaussian maps.
With the rise of 3D foundation models equipped with abundant 3D spatial prior knowledge~\cite{wang2025vggt, leroy2024grounding, smart2024splatt3r}, their SLAM variants effectively leverage this knowledge~\cite{maggio2025vggt, murai2025mast3r, liu2025slam3r} to achieve robust but slightly less precise reconstructions. 


While image blur is a common phenomenon degrading the image quality due to rapid camera motion or scene changes, none of these works explicitly handle blurry inputs, relying only on multi-view optimization and loop closure detection to reduce the impact. 
Recent deblur methods~\cite{bae20242, wang2025mba, peng2024bags, shen2025dof} have achieved good results by explicitly modeling physical motion and defocus blur, and attempting to learn sharp details from blurry inputs. 
However, they have many limitations, \eg the SLAM methods in these works~\cite{bae20242, wang2025mba} mostly focus on simple motion blur, limiting the further development of their performance.

The proposed \methodname{} method aims to further improve localization and high-fidelity reconstruction accuracy in more realistic blur scenarios, while addressing the limitations of the aforementioned methods. It combines explicit and implicit approaches to handle image blur. First, we preliminarily process real blur using the deblur foundation model through our proposed training method. Then, we reduce the impact of blurry frames on the system through multi-view optimization and loop closure detection, and estimate pixel-wise blur kernels through an MLP network and historical information. 
When the deblur network fails, we render multiple blurry sub-frames in 3D space by the MLP network and the global 3DGS~\cite{kerbl3Dgaussians} model. 
Sharp frames will be identified and jointly optimized with other frames for 3DGS reconstruction. As shown in Figure~\ref{fig:teaser}, our system exhibits excellent deblurring effects. We make the following contributions:

\begin{itemize}
\item We propose the first SLAM method capable of simultaneously handling defocus blur and motion blur. Using multi-scale blur kernels estimated via additional MLP layers, after processing by a deblurring neural network, we can handle various  blur types. 
\item We develop an automatic method to detect the amount of blur in the input frame, treat sharp and blurry frames separately, and skip the costly refinement for sharp frames. 
This is in comparison to all previous blur-aware SLAM approaches that assume all input frames to be blurry.
\item We integrate a single-image deblur feed-forward model into a high-fidelity reconstruction SLAM system, achieving 3D consistency through our proposed training method, thereby bringing continuous improvements to both SLAM tracking and reconstruction modules. 
\item We propose a robust online deblurring method based on 3D Gaussians. The method performs initial blur removal based on a feed-forward model and conducts further blur removal and detail enhancement using multi-view blur kernel learning. Meanwhile, the method incorporates loop closures and online Gaussian deformation to further enhance accuracy. The method achieves state-of-the-art performance on multiple datasets. Code will be released.
\end{itemize}

%% file: sec/2_related_works.tex
\section{Related Work}
\label{sec:formatting}

\boldparagraph{Neural Network-Based Deblurring Methods.}
Single-frame image deblurring networks achieve superior restoration quality through problem-tailored architectures. For motion blur tasks, Transformer-based methods including Chen~\etal~\cite{Chen_2025_CVPR} and LoFormer~\cite{mao2024loformer} enhance detail recovery by decomposing motion patterns. However, these 2D deblurring networks do not impose multi-view geometric constraints and are prone to generating 2D noise. Meanwhile, these networks have prohibitively long computation times and cannot perform online inference. In contrast, state space models such as XYScanNet~\cite{liu2025xyscannet} and EVSSM~\cite{kong2025efficientvisualstatespace} adopt Mamba-based architectures to capture long-range dependencies with linear complexity, thereby enhancing real-time performance while maintaining good quality. These networks do not impose physical constraints during training. Furthermore, these networks have not been evaluated on images with high sensor noise~\cite{sturm12iros} and heavy motion blur~\cite{10016760} as common in SLAM settings.

\boldparagraph{Deblurring methods based on 3D Gaussians.}
3D Gaussian Splatting (3DGS)~\cite{kerbl20233d} can effectively utilize historical information to model input image blur in a 3D-consistent manner based on its high-fidelity dense mapping capability. Similar to neural network-based deblurring work, these approaches can be divided into two categories. The first category only models motion blur. For example, BAD-Gaussians~\cite{zhao2024bad} uses historical models to render the physical imaging process of motion blur modeling, recovering rendered sub-frame trajectories while jointly optimizing Gaussian parameters. CoMoGaussian~\cite{lee2025comogaussian} refines the motion modeling by using neural ordinary differential equations to model continuous camera motion trajectories instead of the previous discretized interpolation, and uses per-pixel weights to perform weighted averaging of each sub-frame's contribution to the final rendered blurry image rather than directly averaging the sub-frames. These works often assume that motion is captured under small aperture camera settings with only motion blur. DoF-GS~\cite{wang2025dof} represents another category of works which models defocus blur through a focus localization network. There are also methods that model both types of blur through blur kernels, such as Deblurring 3D Gaussian Splatting~\cite{lee2024deblurring} and BAGS~\cite{peng2024bags}. However, they are limited by kernel size and Gaussian count, failing to handle extreme blur robustly. In contrast, the blur in our system is preliminarily processed by a deblurring neural network.

\boldparagraph{Deblurring SLAM Methods.}
In the SLAM field, advanced theories from offline deblurring 3D reconstruction methods have been leveraged to construct online SLAM approaches with deblurring capabilities. MBA-SLAM~\cite{wang2025mba} adopts the motion blur modeling method based on BAD-Gaussians~\cite{zhao2024bad} for online operation and combines it with a motion blur-aware tracker. I2-SLAM~\cite{bae20242} sets explicit variables for the image formation steps (white balance, exposure time, camera response function) to adapt to appearance changes in each frame, while also using sub-frames to model motion blur. I2-SLAM~\cite{bae20242} attempted to use deblurring neural networks to enhance performance, but failed due to the 3D spatial inconsistency of single-image deblurring networks. Similar to I2-SLAM, Deblur-SLAM~\cite{girlanda2025deblur} further integrates local bundle adjustment and loop closures for improved consistency.
DAGS~\cite{QIN2025113366} builds upon this by adding Gaussian-based deblurring awareness to output binary blur labels, unlike other works that assume all frames are blurred. All these works assume that the camera operates with a small aperture without any defocus blur and they cannot adapt their high computational effort to the amount of blur in the input images.
Since in practice many SLAM datasets contain a mixture of motion blur and defocus blur, as well as various ratios of sharp vs. blurry images, we designed the first method to handle both types of blur and to adapt the computation effort to the estimated amount of blur in the input images for fast computation times.

%% file: sec/3_method.tex
\section{Method}
\label{sec:method}
\methodname{} assumes that real-world blur may have different sources: defocus blur occurs when light forms an out-of-focus image on the camera plane, while motion blur occurs through temporal integration of motion during exposure.
Under this assumption, we propose a physics-constrained method to train a 3D-consistent single-image deblur model to enhance the blur robustness of the Droid-SLAM~\cite{teed2021droid} tracker and the monocular depth estimation network~\cite{eftekhar2021omnidata}.

\begin{figure*}[t]
  \centering
   \includegraphics[width=1.0\linewidth]{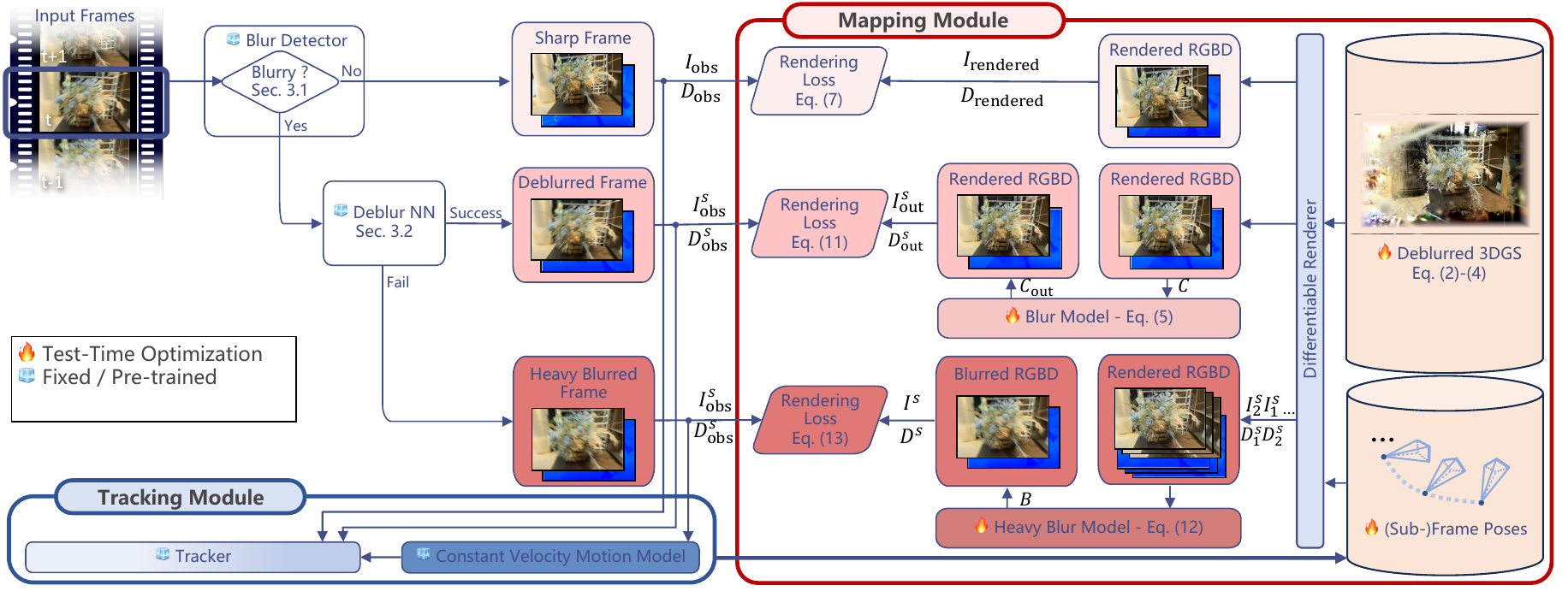}
   \caption{\textbf{Method overview.} \methodname{} robustly handles varying blur by adaptively categorizing images into sharp, blurry, and heavily blurred levels (shown in different red shades). Since typically only a subset of frames is blurred, this improves both blur handling and average runtime. Both tracking and mapping modules optionally leverage the deblurring network.
    The red mapping module optimizes the 3DGS reconstruction using sliding-window (Eq.~\eqref{eq:bundle_adjustment}) and global (Eq.~\eqref{eq:global_bundle_adjustment}) losses, incorporating depth (Eq.~\eqref{eq:update_gaussian}) and pose from the blue tracking module. 
    }
   \label{fig:overview}
\end{figure*}

We classify input frames into three categories based on blur detector scores: sharp frames, blurred frames with failed deblurring, and blurred frames with successful deblurring. Sharp frames are directly processed through Droid-SLAM for pose estimation and depth refinement. The 3D Gaussian representation is further refined using sharp frames through online deformation and rendering optimization guided by the Splat-SLAM~\cite{sandstrom2024splat} loss function.
RGB frames with high blur scores where deblurring fails are skipped by the tracker.
These frames are optimized by calculating photometric loss against motion-blurred frames formed from averaging multiple sub-frames that are blurred through the MLP layer.
For successful deblurring cases, the frame is processed through Droid-SLAM, and the map representation is optimized using Gaussian deformation and photometric and depth errors of the rendered blurry frame modeled by the MLP layer.
As shown in Figure~\ref{fig:overview}, the blur models are divided into two categories: one category uses multiple blur sub-frames to optimize a photometric error with the input image, while the other category uses a single blurry rendered frame to compute the output image from the previous deblur model. 

\subsection{Blur Quantification and Recognition}
\label{sec:blur_quant}
As part of our pipeline, we dynamically detect which frames are sharp and which are blurry. To identify the optimal blur detector for this task, we constructed a comprehensive blur detection benchmark consisting of real-world datasets and two semi-synthetic blur datasets. The real-world data includes RealBlur~\cite{rim_2020_ECCV} for motion blur, as well as DPDD~\cite{abuolaim2020defocus} and RealDoF~\cite{lee2021iterative} for defocus blur. 

To construct the first semi-synthetic dataset, we apply the Eden~\cite{zhang2025eden} video interpolation model to the ScanNet~\cite{dai2017scannet} dataset to generate intermediate frames. We physically model motion blur by averaging these interpolated frames in linear color space, utilizing the middle frame as the ground truth sharp reference. The second semi-synthetic dataset combines sequences from the RED~\cite{Nah_2019_CVPR_Workshops_REDS}, GoPro~\cite{Nah_2017_CVPR}, and ReplicaBlurry~\cite{girlanda2025deblur} datasets, which are manually annotated to remove frames with moving objects or inherent blur.

Using this benchmark, we evaluated 39 different image quality metrics. We assessed them based on a custom proxy measurement that calculates a consistency score derived from the metric's accuracy and effect size. Based on our evaluation, ARNIQA~\cite{agnolucci2024arniqa} achieved the highest consistency score and was selected as our default blur detector. Detailed formulations regarding the blur synthesis and metric evaluations are provided in the supplementary material (Sec.~\ref{sec:supp_blur_quant}).

\subsection{Physics-constrained Deblurring Network}
\label{sec:physics_deblur}
We impose additional physical constraints during the dataset creation process to guide the network in learning 3D-consistent deblurring capabilities. First, according to imaging principles, real-world motion blur is formed when light hits the camera plane and undergoes a certain degree of defocus based on focal length and then accumulates on the exposure plane. In this case, the image forms multiple trailing edges. That is, the image represents a video sequence over the exposure duration. The physical essence of deblurring lies in restoring the image corresponding to a specific moment during exposure. 
Real-world motion blur formation follows the physical process:
\begin{equation}
    \text{B} = \int_{\text{0}}^{\text{T}} \Big( \text{I}\text{(t)} \otimes \text{K}_{\text{defocus}}\big(\text{f}, \text{d(t)}\big) \Big) \mathrm{d}t 
    \label{eq:blur_formation}
\end{equation}
where $\text{B}$ is the blurred image, $\text{I}\text{(t)}$ is the latent sharp image at time $\text{t}$, $\otimes$ denotes the convolution operation, $\text{K}_{\text{defocus}}$ is the defocus kernel, and $T$ is the exposure time.

We select the middle frame as the reconstruction target to avoid ambiguity in motion direction~\cite{pham2023hypercut}. However, the inherent misalignment in real-world motion blur datasets~\cite{rim_2020_ECCV} may lead the model to learn image sharpening instead of accurately extracting the mid-exposure frame. To address this, we adopt a two-stage training strategy: we first train on the RED~\cite{Nah_2019_CVPR_Workshops_REDS}, GoPro~\cite{Nah_2017_CVPR}, and ReplicaBlurry~\cite{girlanda2025deblur} datasets (\ie semi-synthetic datasets) to learn mid-exposure frame extraction. 
We use multiple video frames from real-world and synthetic environments to model motion blur, which allows the model to learn the accurate positions of intermediate frames. The combination of synthetic and real-world data enables it to achieve a certain degree of generalization across different camera models. 
Subsequently, we fine-tune on the DPDD~\cite{abuolaim2020defocus} defocus dataset to enhance robustness against defocus blur. All datasets undergo inverse gamma correction to maintain color constancy.

\subsection{3D Gaussian Splatting Representation}
Following Splat-SLAM~\cite{sandstrom2024splat}, we represent the scene as a collection of 3D Gaussians $\text{G} = \{\text{g}_\text{i}\}$, where each Gaussian $\text{g}_\text{i}$ is parameterized by:
\begin{equation}
\text{g}_\text{i(x)} = \exp\left(-\frac{1}{\text{2}}(\text{x} - \boldsymbol{\mu}_\text{i})^\text{T}\Sigma_\text{i}^{-1}(\text{x} - \boldsymbol{\mu}_\text{i})\right)
\end{equation}
where $\boldsymbol{\mu}_\text{i} \in \text{R}^\text{3}$ is the mean position, $\Sigma_\text{i} \in \text{R}^{\text{3} \times \text{3}}$ is the covariance matrix decomposed as $\Sigma_\text{i} = \text{R}_\text{i} \text{S}_\text{i} \text{S}_\text{i}^\text{T} \text{R}_\text{i}^\text{T}$ with rotation $\text{R}_\text{i}$ and scale $\text{S}_\text{i}$, along with opacity $\text{o}_\text{i} \in [0,1]$ and color $\text{c}_\text{i} \in \text{R}^\text{3}$.
The rendering projects 3D Gaussians onto the image plane and blends them using alpha compositing:
\begin{equation}
\label{eq:projected_gaussian}
\text{C} = \sum_{\text{i} \in \text{N}} \text{c}_\text{i} \alpha_\text{i} \prod_{\text{j}=1}^{\text{i}-1}(1-\alpha_\text{j}), \quad \text{D}_\text{r} = \sum_{\text{i} \in \text{N}} \text{d}_\text{i} \alpha_\text{i} \prod_{\text{j}=1}^{\text{i}-1}(1-\alpha_\text{j})
\end{equation}
where $\alpha_\text{i} = \text{o}_\text{i} \cdot \text{g}_\text{i}^{\text{2D}}(\text{x'})$ represents the combined opacity and 2D Gaussian density at pixel $\text{x'}$, and $\text{d}_\text{i}$ is the depth of Gaussian center.

When Droid-SLAM performs multi-view depth optimization, we accordingly deform the Gaussian representation. Given a depth update from $\text{d}$ to $\text{d'}$ at pixel coordinate $(\text{u}, \text{v})$ in keyframe $\text{K}$, we update the mean position of affected Gaussians:
\begin{equation}
\label{eq:update_gaussian}
\boldsymbol{\mu}_\text{i'} = \boldsymbol{\mu}_\text{i} + \frac{\text{d'} - \text{d}}{\text{d}} (\boldsymbol{\mu}_\text{i} - \text{t}_{\text{K}})
\end{equation}
where $\text{t}_{\text{K}}$ is the camera center of keyframe $\text{K}$. This deformation ensures the Gaussian map remains consistent with the optimized geometry.
\subsection{Complex Blur Modeling}
\label{sec:complex_blur}
For successfully deblurred frames, we adopt the Blur Proposal Network (BPN) from BAGS~\cite{peng2024bags} to model residual blur. BPN estimates per-pixel convolution kernels $\text{h}(\text{x}) \in \text{R}^{\text{K} \times \text{K}}$ and masks $\text{m}(\text{x}) \in [0,1]$:
\begin{equation}
    \label{eq:BAGS}
    \text{C}_{\text{out}}(\text{x}) = \big(1 - \text{m}(\text{x})\big) \text{C}(\text{x}) + \text{m}(\text{x})\big(C(\text{x}) \otimes \text{h}(\text{x})\big)
\end{equation}
where $C$ is the sharp rendered image.
For frames with failed deblurring, we render multiple sub-frames and apply BPN to each, discretizing the temporal integration in Eq.~\eqref{eq:blur_formation}:
\begin{equation}
    \label{eq:bpn_formation}
    \text{B} \approx \frac{1}{\text{N}_{\text{sub}}} \sum_{j=1}^{N_{\text{sub}}} \text{BPN}(\text{I}_\text{j}, \text{h}_\text{j}, \text{m}_\text{j})
\end{equation}

\subsection{Hybrid Bundle Constraint}
\label{sec:hybrid_bundle}
We use deformable Gaussians to update the depth propagated from the front-end tracker during loop closure and multi-frame optimization, while simultaneously updating each camera's pose. 
Meanwhile, frames are categorized as sharp if the blur detection metric~\cite{agnolucci2024arniqa} is below a pre-defined threshold $\tau_{\text{sharp}}$, otherwise as blurry.
For blurry frames, we apply the physics-constrained deblurring network and evaluate success using both Laplacian ratio and blur quantification metric. Successful deblurring is determined when the combined metric exceeds the threshold $\tau_{\text{success}}$.

\boldparagraph{Sharp frame optimization.}
Sharp frames receive higher optimization weights $\text{w}_{\text{sharp}}$ to provide strong priors:
\begin{equation}
\begin{aligned}
\text{L}_{\text{sharp}} &= \text{w}_{\text{sharp}} \bigg[ \lambda_{\text{rgb}} (||\text{I}_{\text{rendered}} - \text{I}_{\text{obs}}||_1) \\
&+ \lambda_{\text{depth}} ||\text{D}_{\text{rendered}} - \text{D}_{\text{obs}}||_1 \bigg]
\end{aligned}
\end{equation}

\boldparagraph{Multi-scale Blurry frame optimization.}
Successfully deblurred frames undergo multi-scale optimization without pose updates. We progressively optimize from coarse to fine scales $\text{s}$ with exposure compensation, where $\text{a}_{\text{exposure}}^{\text{s}}$ and $\text{b}_{\text{exposure}}^{\text{s}}$ are learnable affine scale and shift parameters for the current scale $\text{s}$:
\begin{equation}
\label{eq:scale_loss}
\text{I}_{\text{adjusted}}^{\text{s}}\text{(x)} = \exp(\text{a}_{\text{exposure}}^{\text{s}}) \cdot \text{I}_{\text{rendered}}^{\text{s}}\text{(x)} + \text{b}_{\text{exposure}}^{\text{s}}
\end{equation}
The multi-scale loss combines RGB and depth terms with appropriate masking to handle boundary regions and invalid depth values.
For each scale $\text{s}$, the Blur Proposal Network (BPN) from BAGS~\cite{peng2024bags} learns depth-aware kernels to model residual blur and enhance details:
\begin{equation}
\begin{aligned}
\text{I}_{\text{out}}^\text{s}(\text{x}) = &\big(1 - \text{m}^\text{s}(\text{x})\big) \text{I}_{\text{adjusted}}^\text{s}(\text{x}) \\
&+ \text{m}^\text{s}(\text{x})\big(\text{I}_{\text{adjusted}}^\text{s}(\text{x}) \otimes \text{h}^\text{s}(\text{x})\big)
\end{aligned}
\end{equation}
where $\text{h}^s(\text{x}) \in \text{R}^{\text{K}_\text{s} \times \text{K}_\text{s}}$ is the per-pixel convolution kernel and $\text{m}^\text{s}(\text{x}) \in [0,1]$ is the blending mask.

The kernel $\text{h}^s(\text{x})$ adapts based on the rendered depth $\text{D}(\text{x})$, effectively performing depth-dependent detail restoration:
\begin{equation}
\begin{aligned}
\text{h}^\text{s}\big(\text{x}, \text{D}(\text{x})\big) = \text{h}_{\text{deblur}}^\text{s}\big(\text{D}(\text{x})\big) + \alpha(\text{D}(\text{x})) \!\cdot\! \text{h}_{\text{sharpen}}^\text{s}\big(\text{D}(\text{x})\big) 
\end{aligned}
\end{equation}
where $\text{h}_{\text{deblur}}^\text{s}$ removes residual blur and $\text{h}_{\text{sharpen}}^\text{s}$ enhances edges, with $\alpha(\text{D}(\text{x}))$ being a learnable depth-dependent factor.
For distant regions with larger depth values, the kernel provides stronger enhancement to compensate for angular resolution limitations and depth-dependent degradation. This enhancement remains physically plausible through multi-view consistency in our bundle adjustment.

Unlike BAGS~\cite{peng2024bags}, which requires a warm-up phase, we directly optimize each frame without initialization stages. Rather than employing Mip-Splatting~\cite{Yu2023MipSplattingA3} for multi-resolution consistency, we adopt direct image rescaling. The multi-scale loss combines RGB and depth terms with BPN modeling and appropriate masking to handle boundary regions and invalid depth values:
\begin{equation}
\label{eq:deblur_loss}
\begin{aligned}
\text{L}_{\text{deblur}}^\text{s} = &\lambda_{\text{rgb}} \|\text{I}_{\text{out}}^\text{s} - \text{I}_{\text{obs}}^\text{s}\|_1 + \lambda_{\text{depth}} ||\text{D}_\text{{out}}^\text{s} - \text{D}_{\text{obs}}^\text{s}||_1 \\&+ \lambda_{\text{sparse}} \|\text{m}^\text{s}\|_1 
\end{aligned}
\end{equation}
where $\|\text{m}^{\text{s}}\|_{\text{1}}$ represents the $\text{L}_{\text{1}}$ norm of the mask over all valid pixels $\text{x}$. where the sparsity constraint on $\text{m}^\text{s}$ ensures selective enhancement only where beneficial, preventing overfitting to sensor-specific noise patterns. The $\text{D}_{\text{out}}^\text{s}$ and $\text{I}_{\text{out}}^\text{s}$ denote the depth and image outputs from BPN (Eq.~\eqref{eq:bpn_formation}) at scale $\text{s}$ and the Different from offline reconstruction, we set the mask to 0 for insufficiently reconstructed regions in online reconstruction, preventing overly large or small masks from affecting the inference.
This enhancement improves perceptual quality but may decrease pixel-wise metrics such as PSNR.


\begin{figure*}
  \centering
\includegraphics[width=1.0\linewidth]{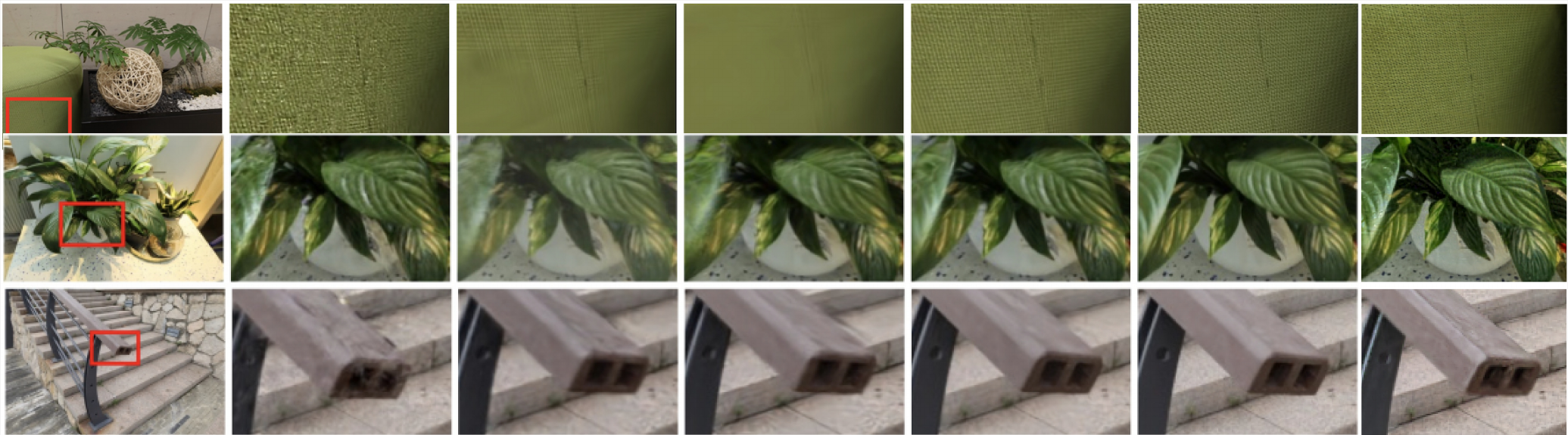}
\vspace{-0.3em}  
Reference \hspace{2.3em} BAD-GS \hspace{2.3em} DeblurringGS \hspace{2.3em} BAGS \hspace{2.3em} CoMoGaussian \hspace{2.3em} Ground Truth \hspace{2.3em}  Ours
\caption{\textbf{Deblurring performance of our method} when compared to other state-of-the-art offline methods.}
   \label{fig:results}
\vspace{-1em}
\end{figure*}

\boldparagraph{Heavy blurry frame optimization.}
For failed deblurring, we employ virtual camera trajectory optimization. We generate $\text{N}_{\text{sub}}$ virtual cameras along the blur trajectory, each parameterized by $(\text{R}_\text{k}, \text{t}_\text{k}, \theta_\text{k}, \rho_\text{k})$. The composite blur model is applied to each rendered sub-frame, and the averaged result is compared with the blurred observation. We evaluate the rendering errors for RGB $\text{I}^\text{s}$ and depth $\text{D}^\text{s}$ images at scale $\text{s}$.

First, we synthesize the blurred approximation $\hat{\text{Q}}^{\text{s}}$:
\begin{equation}
\hat{\text{Q}}^{\text{s}} = \frac{1}{\text{N}_{\text{sub}}} \sum_{\text{j}=1}^{\text{N}_{\text{sub}}} \text{BPN}(\text{Q}_\text{j}^\text{s}, \text{h}_\text{j}^\text{s}, \text{m}_\text{j}^\text{s})
\end{equation}

The failure loss is then calculated as:
\begin{equation}
\label{eq:fail_loss}
\text{L}_{\text{fail}} = \sum_{\text{Q} \in \{\text{I}^\text{s}, \text{D}^\text{s}\}} \lambda_{\text{Q}} \|\hat{\text{Q}}^{\text{s}} - \text{Q}_{\text{obs}}^{\text{s}}\|_{1} + \lambda_{\text{sparse}} \|\text{m}^{\text{s}}\|_{1}
\end{equation}
where $\text{Q}_{\text{j}}^{\text{s}}$ and $\text{Q}_{\text{obs}}^{\text{s}}$ denote the rendered sub-frames (color or depth) and the observed input frame. Other terms are defined in Eq.~\eqref{eq:bpn_formation} and \eqref{eq:deblur_loss}.

\boldparagraph{Bundle adjustment.}
Joint bundle optimization is performed across all frame types with adaptive weighting:
\begin{equation}
\label{eq:bundle_adjustment}
\text{L}_{\text{total}} = \sum_{\text{f} \in \text{F}} \text{w}_\text{f} \text{L}_\text{f}
\end{equation}
where $\text{L}_{\text{f}}$ is the frame-specific loss term, which equals $\text{L}_{\text{sharp}}$ for sharp frames, or $\text{L}_{\text{deblur}}^{\text{s}}$ / $\text{L}_{\text{fail}}$ for blurry frames.
The $\text{w}_{\text{sharp}} > \text{w}_{\text{deblur}} = \text{w}_{\text{fail}}$ to prioritize sharp observations. Unlike Splat-SLAM's~\cite{sandstrom2024splat} selective keyframe filtering, we incorporate all tracker keyframes as backend optimization keyframes to fully leverage deblurring results.

\subsection{Global Optimization and Loop Closure}
Following Splat-SLAM~\cite{sandstrom2024splat}, we maintain global consistency through the following three main paradigms.

\boldparagraph{$\triangleright$ Local bundle adjustment} optimizes within a sliding window of recent keyframes using the Disparity, Scale, and Pose Optimization (DSPO) layer.

\boldparagraph{$\triangleright$ Loop closure detection} identifies revisited locations through optical flow magnitude thresholding and temporal constraints.

\boldparagraph{$\triangleright$ Global bundle adjustment} periodically optimizes the entire pose graph with appropriately deformed Gaussians:
\begin{equation}
\label{eq:global_bundle_adjustment}
\text{L}_{\text{global}} = \sum_{\text{f} \in \text{F}} \text{w}_\text{f} \text{L}_\text{f} + \lambda_{\text{reg}} \sum_{\text{i}=\text{1}}^{\text{N}} \|\text{s}_\text{i} - \bar{\text{s}}\|_\text{1}
\end{equation}
where $\text{w}_\text{f}$ are frame-specific weights prioritizing sharp observations, and the regularization term with $\lambda_{\text{reg}}>0$ prevents largely elongated Gaussians.

\subsection{Final Refinement}
Upon trajectory completion, we perform final optimization with multi-scale progressive refinement, re-estimating poses and exposure parameters for failed deblurring frames while maintaining global consistency for all other frames.

%% file: sec/4_experiments.tex
\section{Experiments}
\label{sec:experiment}

\begin{figure}[t]
  \centering
  \setlength{\tabcolsep}{2pt}
  \begin{tabular}{cc}
    \includegraphics[width=0.46\linewidth]{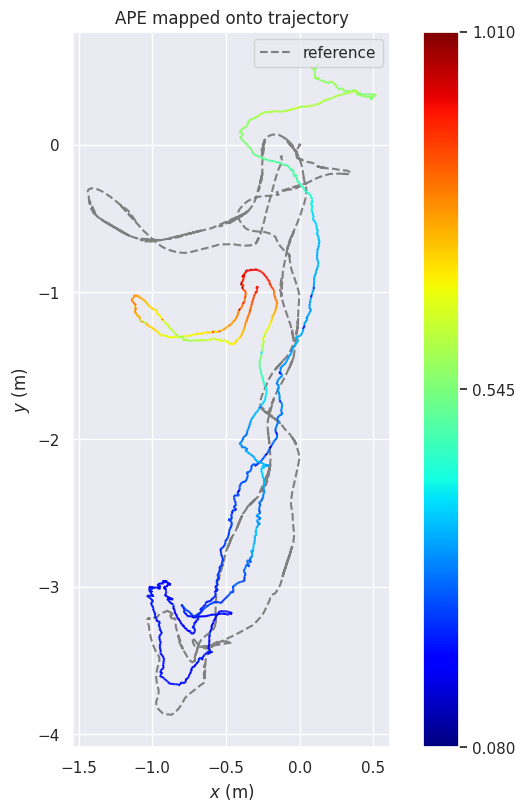} &
    \includegraphics[width=0.48\linewidth]{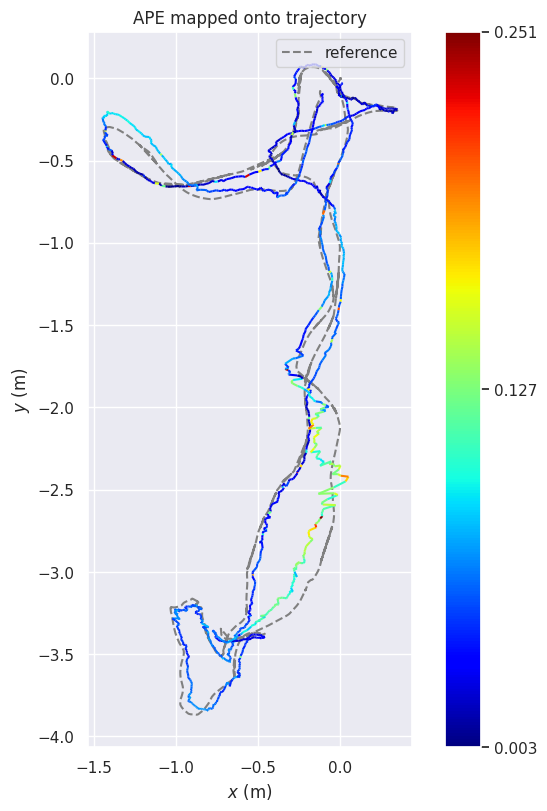} \\
    (a) Droid-SLAM & (b) Ours
  \end{tabular}
  \vspace{-0.5em}
  \caption{\textbf{Trajectory comparison with Droid-SLAM} on the indoor MCD dataset~\cite{10016760}.}
  \label{fig:ate_qualitative}
\end{figure}

\boldparagraph{Datasets.}
To evaluate our method under extreme motion blur conditions, we use a synthetically blurred ReplicaBlurry~\cite{girlanda2025deblur} dataset generated by (Sec.~\ref{sec:blur_quant}) and ArchViz dataset generated by the Unreal game engine\cite{liu2021mba}. We also test our method on real-world TUM-RGBD~\cite{sturm12iros} and IndoorMCD~\cite{10016760} datasets to assess performance under different camera conditions with handheld devices and wheeled platforms. 
To measure the deblurring quality of our method, we further evaluate using the Deblur-NeRF~\cite{ma2022deblur} dataset, which is commonly used for offline deblurring methods.

\boldparagraph{Metrics.}
For experiments on synthetic datasets, we report PSNR between rendered keyframe images and the original sharp images to demonstrate the robustness of our method in extreme cases. 
For the Deblur-NeRF~\cite{ma2022deblur} dataset, we report PSNR, SSIM~\cite{1284395}, and LPIPS~\cite{zhang2018unreasonable} metrics with ground truth sharp images.
For other real-world datasets (TUM-RGBD~\cite{sturm12iros}, IndoorMCD~\cite{10016760}), we only report ATE[m] due to the lack of sharp ground truth frames.
Further, we evaluated PSNR on 3 sequences from the TUM dataset with manually annotated sharp frames provided by I2-SLAM. 

\boldparagraph{Implementation details.}
We set three scales for mapping optimization at 3, 5, and 9. Under severe blur, we model motion blur using 3 virtual sub-frames. Experiments are conducted on an AMD EPYC-2 7282 processor with an NVIDIA RTX A6000 equipped with 48GB memory. 

\begin{table*}[t]
\centering
\small
\setlength{\tabcolsep}{15pt} 
\newcommand{\myskip}{\hskip 3em} 
\begin{tabular}{l@{\myskip}cc@{}c@{\myskip}cc@{}c@{\myskip}cc@{}c@{\myskip}c}
\toprule
\multirow{2}{*}{Method} & \multicolumn{2}{c}{ArchViz-1} & & \multicolumn{2}{c}{ArchViz-2} & & \multicolumn{2}{c}{ArchViz-3} & & \multirow{2}{*}{Avg. ATE} \\
\cmidrule(lr){2-3} \cmidrule(lr){5-6} \cmidrule(lr){8-9} 
& ATE $\downarrow$ & PSNR $\uparrow$ & & ATE $\downarrow$ & PSNR $\uparrow$ & & ATE $\downarrow$ & PSNR $\uparrow$ & & \\
\midrule
MBA-SLAM~\cite{wang2025mba} & \textbf{0.0075} & 28.45 & & 0.0036 & 30.16 & & 0.0141 & 27.85 & & 0.0084 \\
Ours & \textbf{0.0075} & \textbf{28.76} & & \textbf{0.0027} & \textbf{32.71} &  & \textbf{0.0067} & \textbf{30.09} & & \textbf{0.0056} \\
\bottomrule
\end{tabular}
\caption{\textbf{Qualitative experiments on the ArchViz}~\cite{wang2025mba} synthetic dataset to demonstrate the performance comparison between our method and MBA-SLAM~\cite{wang2025mba}. This dataset contains extreme blur cases, and we achieve the best results using ATE RMSE [m] and PSNR metrics.}
\label{tab:archviz_comparison}
\vspace{-0.1cm}
\end{table*}

\begin{table*}
\centering
\small
\setlength{\tabcolsep}{12.9pt}
\begin{tabular}{lcccccc}
\toprule
{Method} & Deblur-NeRF~\cite{ma2022deblur} & DP-NeRF~\cite{lee2023dp} & BAGS~\cite{peng2024bags} & Deblur-GS~\cite{lee2024deblurring} & DOF-GS~\cite{wang2025dof} & {Ours} \\
\midrule
PSNR $\uparrow$  & 23.98 & 24.12 & 24.17 & 24.21 & 24.12 & \textbf{27.45} \\
\bottomrule
\end{tabular}
\caption{\textbf{Comparison on the Deblur-NeRF defocus blur} dataset~\cite{ma2022deblur}. We report PSNR (dB) $\uparrow$ metric and compare to several state-of-the-art methods. Our method outpeforms other methods by a wide margin, both quantitatively and qualitatively.}
\label{tab:defocusblur}
\vspace{-0.3cm}
\end{table*}

\boldparagraph{Evaluation.}
We first conduct a qualitative evaluation of the blur detector's performance on synthetic and real blurred datasets, as described in Sec.~\ref{sec:blur_quant}. We select the metric with the highest consistency score as our blur detector. We then evaluate pose estimation performance in real-world environments, as shown in Table~\ref{tab:ate_mean}. Ablation studies on different training datasets validate the correctness of our theory. To demonstrate whether our model learns 3D-consistent physical properties, we apply the deblurring model (without blur detector filtering) to the Droid-SLAM pipeline. To assess performance across diverse scenarios, we evaluate our system using 19 sequences from the TUM RGB-D dataset and 57 sequences from the MCD dataset. As shown in Figure \ref{fig:ate_qualitative}, we get high ATE improvement.

To further evaluate our method's robustness against severe motion blur, we conduct an ablation study on the proposed fallback mechanism using the synthetic ReplicaBlurry dataset. Please refer to Sec.~\ref{sec:supp_ablation} in the supplementary material for detailed experimental settings and quantitative results. We further tested our method on the synthetic dataset ArchViz, as shown in Table~\ref{tab:archviz_comparison}. We are more robust than MBA-SLAM on datasets where all frames are blurred.

\begin{table}
\centering
\setlength{\tabcolsep}{0.6em} 
\begin{tabular}{lcccc}
\toprule
{Dataset} & {Droid-SLAM~\cite{teed2021droid}} & {Real\_J} & {Ours*} & {Ours} \\
\midrule
TUM & 0.380 & 0.400 & 0.352 & \textbf{0.336} \\
MCD & 0.138 & 0.139 & 0.155 & \textbf{0.128} \\
\bottomrule
\end{tabular}
\caption{\textbf{Comparison of trajectory errors} ATE RMSE [m] $\downarrow$.
Our method yields the best tracking results. 
Real\_J differs from Ours by replacing the DPDD dataset with the RealBlur\_J dataset. 'Ours' differs from 'Ours*' by training on five ReplicaBlurry~\cite{girlanda2025deblur} sequences instead of one.}
\label{tab:ate_mean}
\end{table}

We also tested our method's performance in mapping. After communicating with the authors of I2-SLAM~\cite{bae20242}, we obtained their manually annotated sharp frames on the TUM dataset and the keyframes used by their system. Under the same keyframes, we compared our system with I2-SLAM. As shown in Table~\ref{tab:tum_psnr}, our system achieved state-of-the-art performance in regions with a high proportion of blurred frames. As show in Figure \ref{fig:mba}, we also have a better result than MBA-SLAM. I2-SLAM also attempted to use the single-image deblurring model to improve their method's effectiveness, but it failed due to the model's 3D inconsistency. Our method, after further incorporation of Gaussian-based deblurring, demonstrated detail enhancement that surpasses annotated sharp images~\cite{bae20242,ma2022deblur} as shown in Figures~\ref{fig:defocus_com} and~\ref{fig:tum_com_gt}. 
This enhancement disrupts the evaluation of reference-based mapping metrics that assume a perfect ground truth.
This phenomenon often occurs in large scenes when the camera moves forward and backward, causing the same object to occupy inconsistent pixel sizes across multiple views. In small scenes, such as the Deblur-NeRF motion blur dataset, our complete system shows significant improvement over the system without Gaussian-based deblurring, as shown in Table~\ref{tab:motionblur} and Figure~\ref{fig:results}. 
In addition, we conducted experiments to observe the performance of using our method in the Deblur-NeRF defocus blur dataset, as shown in Table~\ref{tab:defocusblur}. We compare to offline methods on the two subsets of the Deblur-NeRF dataset.

\begin{table}
\centering
\setlength{\tabcolsep}{7.4pt} 
\begin{tabular}{lccc}
\toprule
Method       & PSNR$\uparrow$ & SSIM$\uparrow$ & LPIPS$\downarrow$ \\
\midrule
Mip-Splatting~\cite{Yu2023MipSplattingA3}  & 21.87 & 0.6270 & 0.3066 \\
Deblur-NeRF~\cite{ma2022deblur}            & 25.63 & 0.7645 & 0.1820 \\
BAD-Gaussians~\cite{zhao2024bad}           & 21.69 & 0.6471 & 0.1262 \\
Deblurring 3DGS~\cite{lee2024deblurring}   & 26.61 & 0.8224 & 0.1096 \\
BAGS~\cite{peng2024bags}                   & 26.70 & 0.8237 & 0.0956 \\
CoMoGaussian~\cite{lee2025comogaussian}    & 27.85 & 0.8431 & 0.0822 \\
\midrule
Ours w/o ref.                              & 28.22 & 0.9007 & 0.1053 \\
Ours                                       & \textbf{29.49} & \textbf{0.9213} & \textbf{0.0728} \\
\bottomrule
\end{tabular}
\caption{\textbf{Comparison against offline deblurring methods} on the Deblur-NeRF dataset~\cite{ma2022deblur}. The comparison on both the motion blur and the defocus blur dataset of this established benchmark for offline deblurring methods demonstrates that \methodname{} is able to outperform current state-of-the-art despite solving the harder online problem in which only partial data is available.}
\label{tab:motionblur}
\end{table}

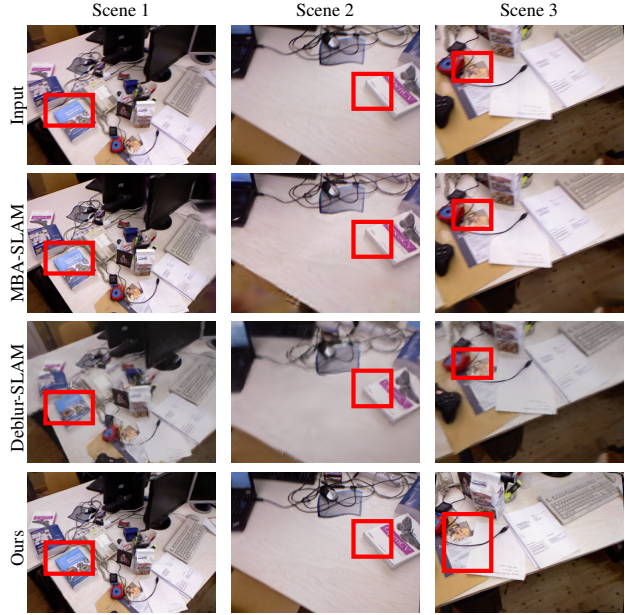
\begin{figure}[t]
  \centering
  \setlength{\tabcolsep}{3pt}
  \newcommand{\imgwidth}{0.3}  
  \begin{tabular}{l@{}ccc}
    & \scriptsize Scene 1 & \scriptsize Scene 2 & \scriptsize Scene 3 \\
    %
    \raisebox{\height}{\rotatebox{90}{\scriptsize Input}} &
    \begin{tikzpicture}
      \node[anchor=south west,inner sep=0] (img) at (0,0) {\includegraphics[width=\imgwidth\linewidth]{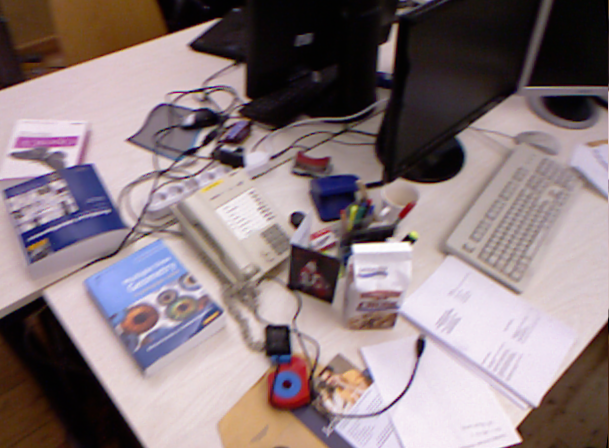}};
      \begin{scope}[x={(img.south east)},y={(img.north west)}]
        \draw[red,ultra thick] (0.10,0.28) rectangle (0.35,0.5);
      \end{scope}
    \end{tikzpicture} &
    \begin{tikzpicture}
      \node[anchor=south west,inner sep=0] (img) at (0,0) {\includegraphics[width=\imgwidth\linewidth]{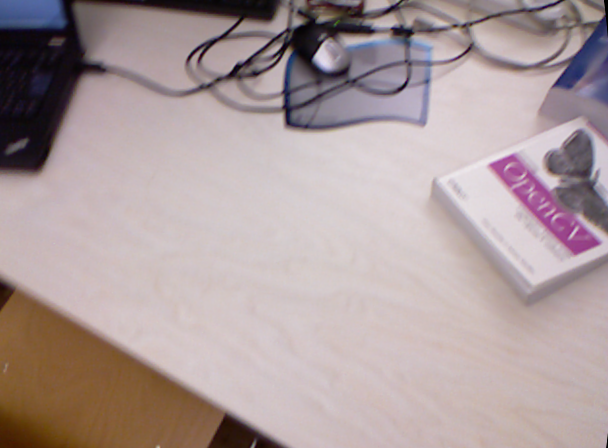}};
      \begin{scope}[x={(img.south east)},y={(img.north west)}]
        \draw[red,ultra thick] (0.65,0.4) rectangle (0.85,0.65);
      \end{scope}
    \end{tikzpicture} &
    \begin{tikzpicture}
      \node[anchor=south west,inner sep=0] (img) at (0,0) {\includegraphics[width=\imgwidth\linewidth]{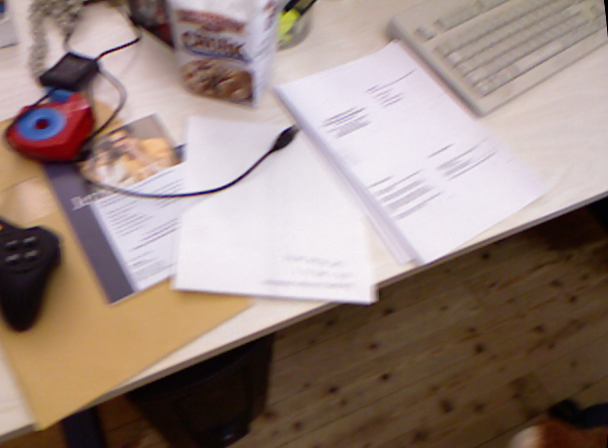}};
    \begin{scope}[x={(img.south east)},y={(img.north west)}]
        \draw[red,ultra thick] (0.1,0.6) rectangle (0.3,0.8);
      \end{scope}
    \end{tikzpicture} \\
    %
    \raisebox{+.1\height}{\rotatebox{90}{\scriptsize MBA-SLAM}} &
    \begin{tikzpicture}
      \node[anchor=south west,inner sep=0] (img) at (0,0) {\includegraphics[width=\imgwidth\linewidth]{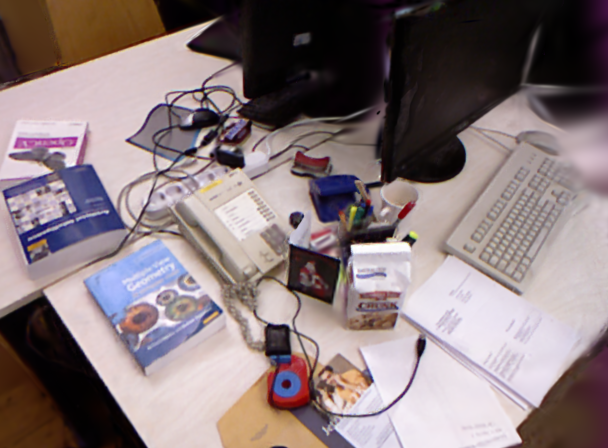}};
      \begin{scope}[x={(img.south east)},y={(img.north west)}]
        \draw[red,ultra thick] (0.10,0.28) rectangle (0.35,0.5);
      \end{scope}
    \end{tikzpicture} &
    \begin{tikzpicture}
      \node[anchor=south west,inner sep=0] (img) at (0,0) {\includegraphics[width=\imgwidth\linewidth]{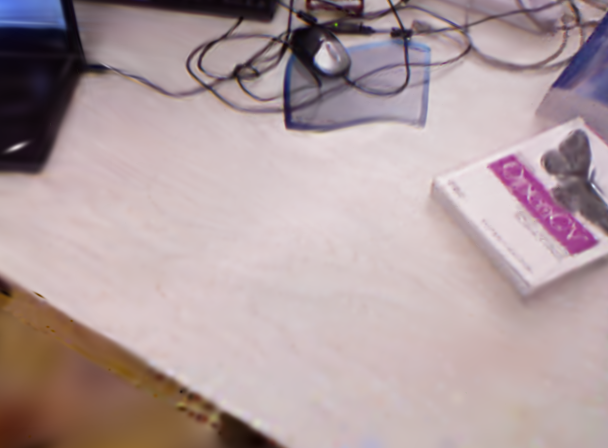}};
      \begin{scope}[x={(img.south east)},y={(img.north west)}]
        \draw[red,ultra thick] (0.65,0.4) rectangle (0.85,0.65);
      \end{scope}
    \end{tikzpicture} &
    \begin{tikzpicture}
      \node[anchor=south west,inner sep=0] (img) at (0,0) {\includegraphics[width=\imgwidth\linewidth]{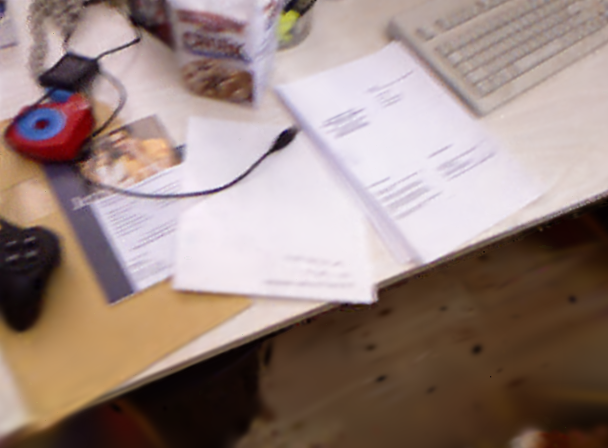}};
      \begin{scope}[x={(img.south east)},y={(img.north west)}]
        \draw[red,ultra thick] (0.1,0.6) rectangle (0.3,0.8);
      \end{scope}
    \end{tikzpicture} \\
    %
    \raisebox{+.1\height}{\rotatebox{90}{\scriptsize Deblur-SLAM}} &
    \begin{tikzpicture}
      \node[anchor=south west,inner sep=0] (img) at (0,0) {\includegraphics[width=\imgwidth\linewidth]{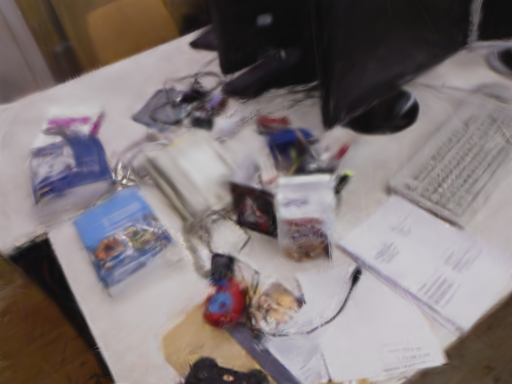}};
      \begin{scope}[x={(img.south east)},y={(img.north west)}]
        \draw[red,ultra thick] (0.10,0.28) rectangle (0.35,0.5);
      \end{scope}
    \end{tikzpicture} &
    \begin{tikzpicture}
      \node[anchor=south west,inner sep=0] (img) at (0,0) {\includegraphics[width=\imgwidth\linewidth]{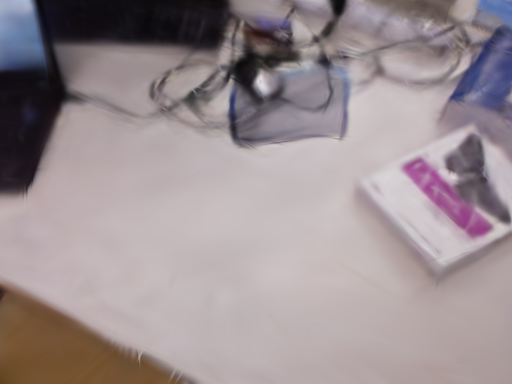}};
      \begin{scope}[x={(img.south east)},y={(img.north west)}]
        \draw[red,ultra thick] (0.65,0.4) rectangle (0.85,0.65);
      \end{scope}
    \end{tikzpicture} &
    \begin{tikzpicture}
      \node[anchor=south west,inner sep=0] (img) at (0,0) {\includegraphics[width=\imgwidth\linewidth]{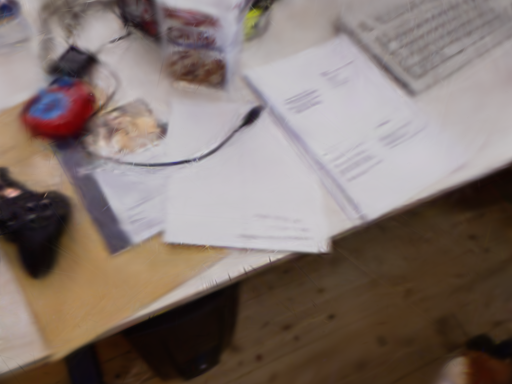}};
      \begin{scope}[x={(img.south east)},y={(img.north west)}]
        \draw[red,ultra thick] (0.1,0.6) rectangle (0.3,0.8);
      \end{scope}
    \end{tikzpicture} \\
    %
    \raisebox{1.3\height}{\rotatebox{90}{\scriptsize Ours}} &
    \begin{tikzpicture}
      \node[anchor=south west,inner sep=0] (img) at (0,0) {\includegraphics[width=\imgwidth\linewidth]{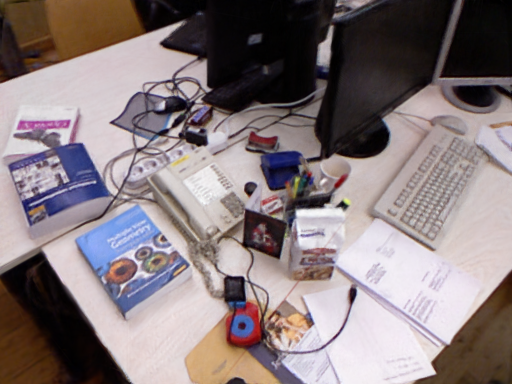}};
      \begin{scope}[x={(img.south east)},y={(img.north west)}]
        \draw[red,ultra thick] (0.10,0.28) rectangle (0.35,0.5);
      \end{scope}
    \end{tikzpicture} &
    \begin{tikzpicture}
      \node[anchor=south west,inner sep=0] (img) at (0,0) {\includegraphics[width=\imgwidth\linewidth]{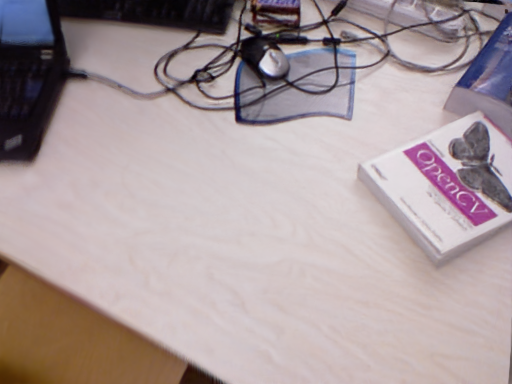}};
      \begin{scope}[x={(img.south east)},y={(img.north west)}]
        \draw[red,ultra thick] (0.65,0.4) rectangle (0.85,0.65);
      \end{scope}
    \end{tikzpicture} &
    \begin{tikzpicture}
      \node[anchor=south west,inner sep=0] (img) at (0,0) {\includegraphics[width=\imgwidth\linewidth]{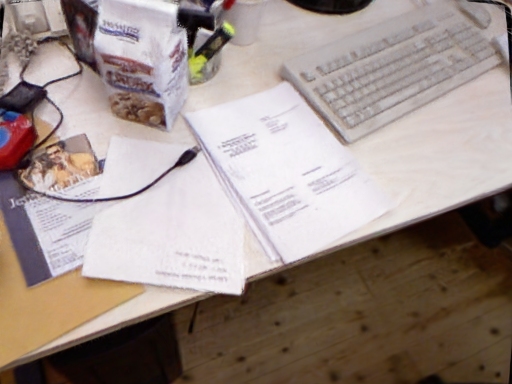}};
      \begin{scope}[x={(img.south east)},y={(img.north west)}]
        \draw[red,ultra thick] (0.05,0.3) rectangle (0.3,0.7);
      \end{scope}
    \end{tikzpicture}
  \end{tabular}
  \vspace{-5pt}
  \caption{\textbf{Qualitative comparison with MBA-SLAM}~\cite{wang2025mba} on the TUM dataset~\cite{sturm12iros}. Our method yields sharper reconstruction results in many scene parts. The qualitative experimental results for the shown fr1\_desk sequence were obtained through communication with the MBA-SLAM authors.}
  \label{fig:mba}
\end{figure}

\begin{table}
\small
\centering
\setlength{\tabcolsep}{0.8em} 
\begin{tabular}{lccc}
\toprule
 & {fr1\_desk} & {fr2\_xyz} & {fr3\_office} \\ 
\midrule
{Sharp ratio} & {16\%} & {100\%} & {37\%}\\
\midrule
I$^2$-SLAM~\cite{bae20242} & 27.23 & \textbf{32.06} & 28.91 \\
Deblur-SLAM~\cite{girlanda2025deblur} & 19.93 & 27.21 & 21.67 \\
Ours   & \textbf{28.03} & 31.14 & \textbf{29.22} \\ 
\bottomrule
\end{tabular}
\caption{\textbf{Rendering comparison (PSNR)} on TUM-RGBD dataset~\cite{sturm12iros}. For fair comparison, we only use the same keyframes as in I2-SLAM~\cite{bae20242} to obtain results, which are then compared against the sharp frames of the keyframes marked by I2-SLAM. The sharp ratio indicates the percentage of sharp keyframes in the input sequence.}
\label{tab:tum_psnr}
\end{table}

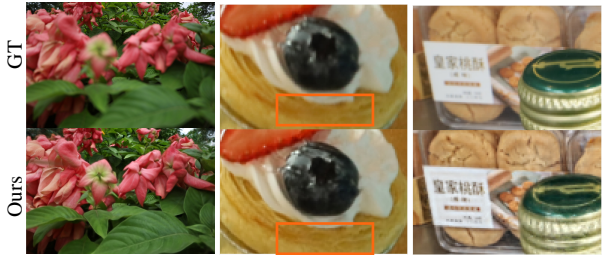
\begin{figure}[t]
\centering
{\footnotesize
\setlength{\tabcolsep}{1pt}
\renewcommand{\arraystretch}{1}
\newcommand{\sz}{0.30}
\begin{tabular}{cccc}
& \includegraphics[valign=b,width=\sz\linewidth]{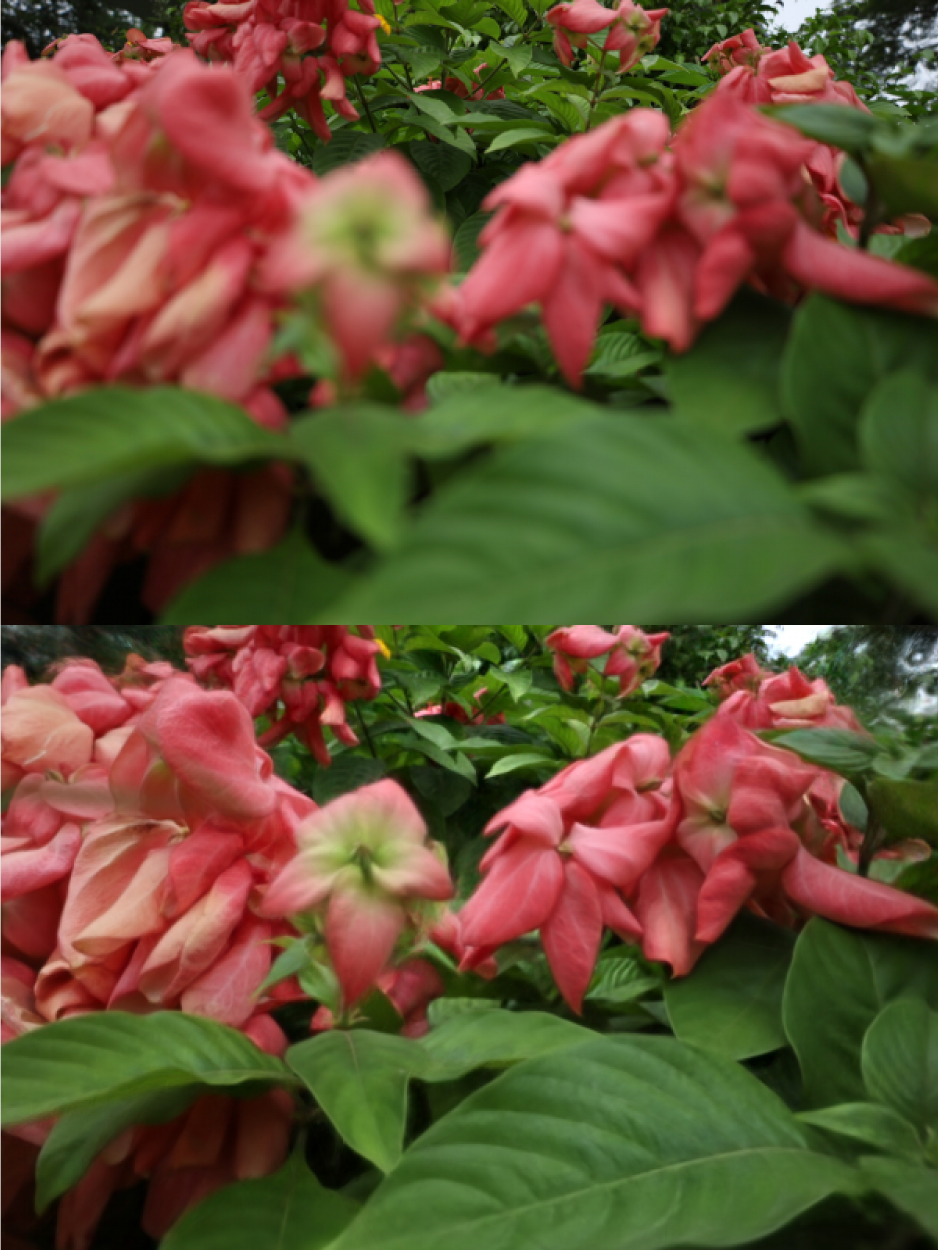} &
\includegraphics[valign=b,width=\sz\linewidth]{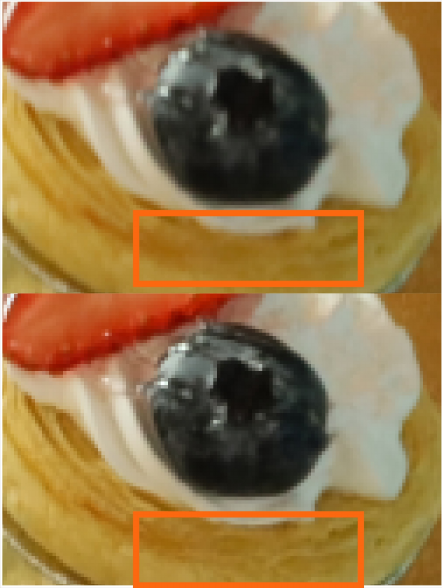} & 
\includegraphics[valign=b,width=\sz\linewidth]{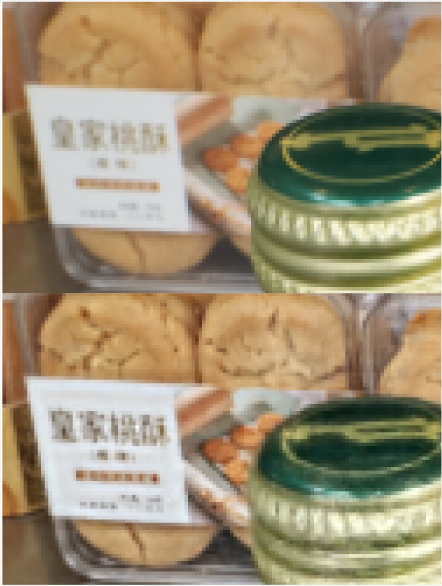}
\\
\end{tabular}
\begin{tikzpicture}[overlay, remember picture]
    \node[rotate=90, anchor=west] at (-7.9cm, 0.8cm) {GT};
    \node[rotate=90, anchor=north] at (-8.1cm, -0.8cm) {Ours};
\end{tikzpicture}
}
\caption{\textbf{Qualitative comparison of our method} with ground truth sharp frames (top row) from the Deblur-NeRF defocus dataset~\cite{ma2022deblur}.}
\label{fig:defocus_com}
\end{figure}

\begin{figure}[t]
\centering
{\footnotesize
\setlength{\tabcolsep}{1pt}
\renewcommand{\arraystretch}{1}
\newcommand{\sz}{0.30}
\begin{tabular}{cccccc}
\rotatebox[origin=c]{90}{} & 
\rotatebox[origin=c]{90}{} & 
\includegraphics[valign=b,width=\sz\linewidth]{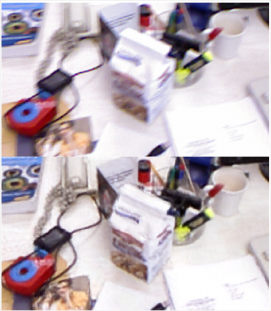} &
\includegraphics[valign=b,width=\sz\linewidth]{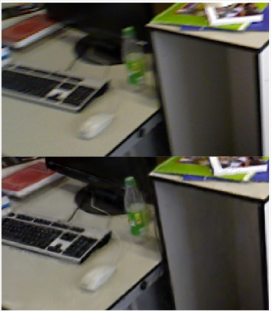} & 
\includegraphics[valign=b,width=\sz\linewidth]{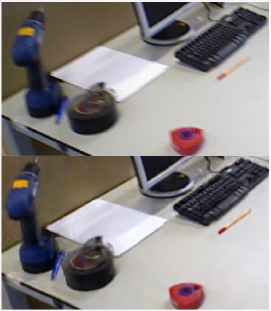}  \\
\end{tabular}

\begin{tikzpicture}[overlay, remember picture]
    \node[rotate=90, anchor=west] at (-3.9cm, 1.90cm) {GT};
    \node[rotate=90, anchor=north] at (-4.1cm, 0.75cm) {Ours};
\end{tikzpicture}
}

\caption{\textbf{Qualitative comparison results on sharp frames manually labeled in I2-SLAM\cite{bae20242}}. The first row is the ground truth image, the second row is the experimental results of our system.}
\label{fig:tum_com_gt}
\end{figure}
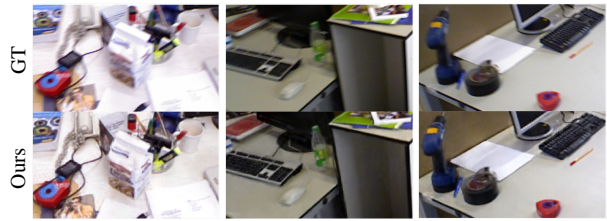

\boldparagraph{Runtime.} 
We used three sequences from the TUM dataset: fr1\_desk, fr2\_xyz, and fr3\_office to perform runtime analysis of our method under online settings in Table~\ref{tab:fps_comparison}.
We also ablated another version of our method without the Gaussian refinement-based approach using only the single-image deblurring neural network. It can be observed that our method achieves excellent real-time performance, and the limitation in real-time performance is primarily due to the deblur neural network which performs inference at 600ms per frame on our device.

\begin{table}
\centering
\setlength{\tabcolsep}{4pt}
\begin{tabular}{lcccc}
\toprule
Method & \textcolor{gray}{Splat-SLAM} & $I^2$-SLAM & Ours &  Ours w/o ref. \\
\midrule
FPS$\uparrow$ & \textcolor{gray}{1.24} & 0.095 & 0.74  & \textbf{0.85} \\
\bottomrule
\end{tabular}
\caption{\textbf{Runtime comparison in frames per second (FPS)}. Ours w/o ref. indicates that we only use the deblurring model without refinement, while "Ours" refers to our complete pipeline. We compute the average FPS of our system on the TUM dataset sequences: fr1\_desk, fr2\_xyz, and fr3\_office.
Note that Splat-SLAM is not a deblurring method and is only listed for reference.}
\label{tab:fps_comparison}
\end{table}

%% file: sec/5_conclusion.tex
\section{Conclusions}

We propose Unblur-SLAM, a novel online deblurring approach that is able to handle both motion and defocus blur to reconstruct sharp 3D scenes from blurred images. By applying middle frame constraints and assumptions about the blur process, we enable single-image deblurring large models to learn 3D-consistent deblurring priors.
Based on the physical assumptions of the blur formation, we construct a Gaussian-based system to estimate the blur kernels of the input images.
To prevent system failure in extreme cases, we propose directly rendering multiple blurred sub-frames in 3D to model severe blur scenarios. Experiments on several real-world datasets show that \methodname{} surpasses state-of-the-art methods in 3D scene deblurring. Unfortunately, our current system remains a trade-off between real-time efficiency and accuracy, requiring further improvements in runtime and memory consumption.

%% file: sec/X_suppl.tex
\clearpage
\setcounter{page}{1}
\maketitlesupplementary

This supplementary material provides additional experimental results and further details on the experiments and methodology. The accompanying video further showcases experimental results.

\section{Experiment Details}
\label{sec:experiment_details}

\subsection{Ablation of the Fallback Mechanism}
\label{sec:supp_ablation}

To demonstrate the effectiveness and necessity of our fallback mechanism under severe motion blur, we conduct an ablation study on the ReplicaBlurry dataset. This dataset simulates severe motion blur by averaging 36 Replica frames. We specifically select the sequence with the largest variation in motion blur degree for this evaluation.

As shown in Table~\ref{tab:fallback_ablation}, the inclusion of the fallback mechanism improves the reconstruction quality, raising the PSNR from 29.38 dB to 29.94 dB, demonstrating its critical role in robustly handling extreme blur cases.

\begin{table}[h]
    \centering
    \begin{tabular}{lc}
        \toprule
        Scenario & PSNR [dB] \\
        \midrule
        With Fallback Mechanism & \textbf{29.94} \\
        Without Fallback Mechanism & 29.38 \\
        \bottomrule
    \end{tabular}
    \caption{\textbf{Ablation of the fallback mechanism.} ReplicaBlurry dataset simulates severe motion blur by averaging 36 Replica frames. We select the sequence with the largest variation in motion blur degree to demonstrate the effectiveness and necessity of our fallback mechanism.}
    \label{tab:fallback_ablation}
\end{table}
\subsection{Additional Information}
\label{sec:add_info}
In the qualitative experiments in Fig.~\ref{fig:mba}, we run Deblur Gaussian SLAM~\cite{girlanda2025deblur} to compare the performance of its system with our system in TUM dataset~\cite{sturm12iros}. It can be seen that, compared to methods based on multiple sub-frames, it is not precise for mild blurring and easily falls into local minima. Especially in the online setting, the rendered images inherently have a certain degree of blurriness due to insufficient observations.

The synthetic dataset ArchViz~\cite{wang2025mba} in Table~\ref{tab:archviz_comparison} was obtained through communication with the MBA-SLAM authors. For fair comparison and reasonable blur modeling, we use a deblurring neural network that was not trained on defocus datasets for preliminary blur removal. Meanwhile, to avoid introducing overfitting to synthetic data, we only used one sequence from the synthetic dataset. Specifically, this neural network was trained only on the Red~\cite{Nah_2019_CVPR_Workshops_REDS}, GoPro~\cite{Nah_2017_CVPR}, and Office3 sequences (from the ReplicaBlurry~\cite{girlanda2025deblur} dataset) after manually filtering out frames containing moving objects.

In Figure~\ref{fig:ate_qualitative}, we demonstrate the ATE performance on the MCD~\cite{10016760} s2r16 sequence to show that under our proposed training method, the single-image deblurring neural network can achieve good 3D consistency, thereby significantly improving multi-view pose estimation performance. We present the trajectory plots of Droid-SLAM~\cite{teed2021droid} and our method combined with Droid-SLAM.

\begin{figure}[tb]
    \centering
    \small
    \setlength{\tabcolsep}{1pt} 
    \renewcommand{\arraystretch}{0.8} 
    \newcommand{\sz}{0.15}
    \begin{tabular}{c ccc}
        
        \rotatebox{90}{\hspace{10pt}Sample 1} &
        \includegraphics[width=\sz\textwidth]{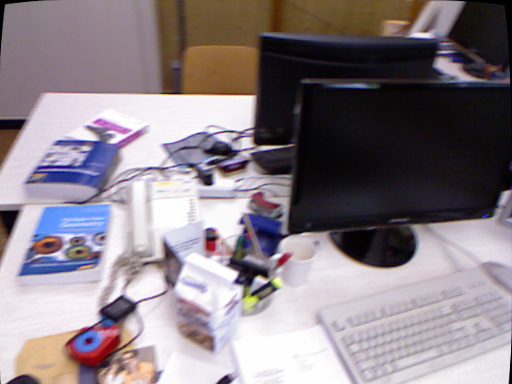} &
        \includegraphics[width=\sz\textwidth]{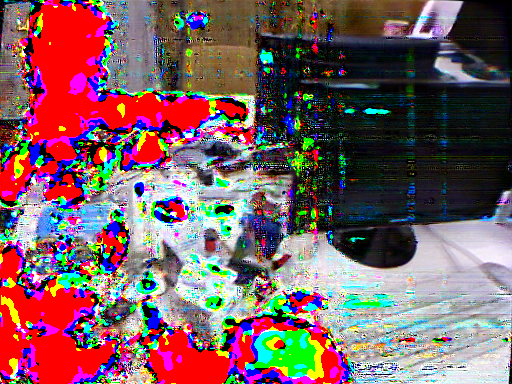} &
        \includegraphics[width=\sz\textwidth]{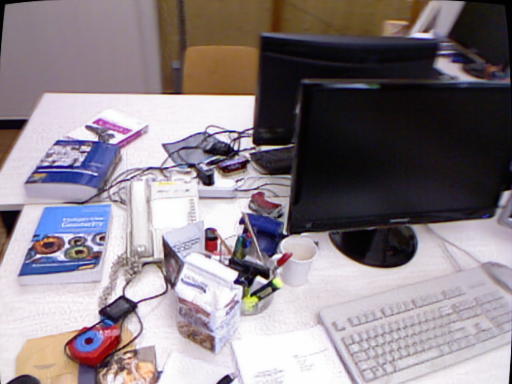} \\
        
        \rotatebox{90}{\hspace{10pt}Sample 2} &
        \includegraphics[width=\sz\textwidth]{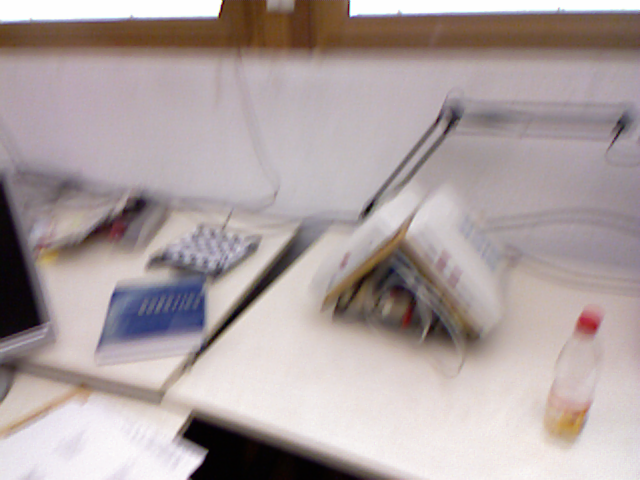} &
        \includegraphics[width=\sz\textwidth]{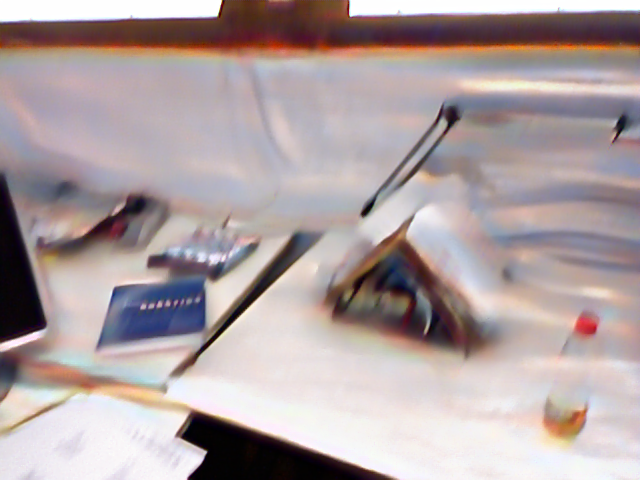} &
        \includegraphics[width=\sz\textwidth]{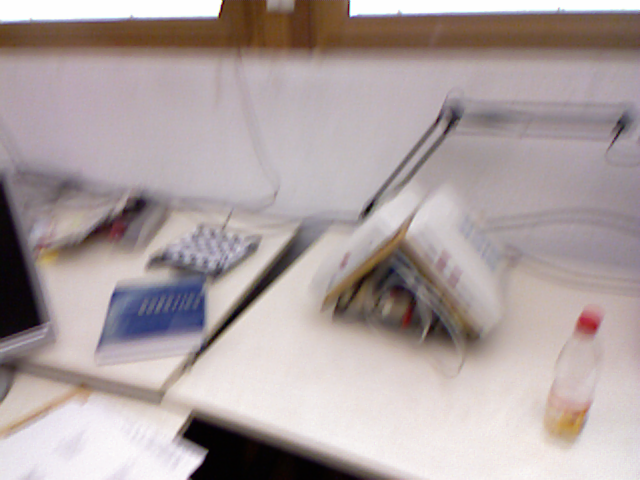} \\
        
        \rotatebox{90}{\hspace{10pt}Sample 3} &
        \includegraphics[width=\sz\textwidth]{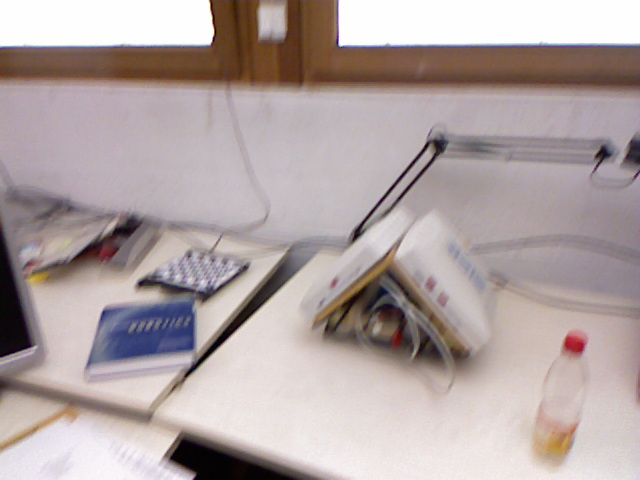} &
        \includegraphics[width=\sz\textwidth]{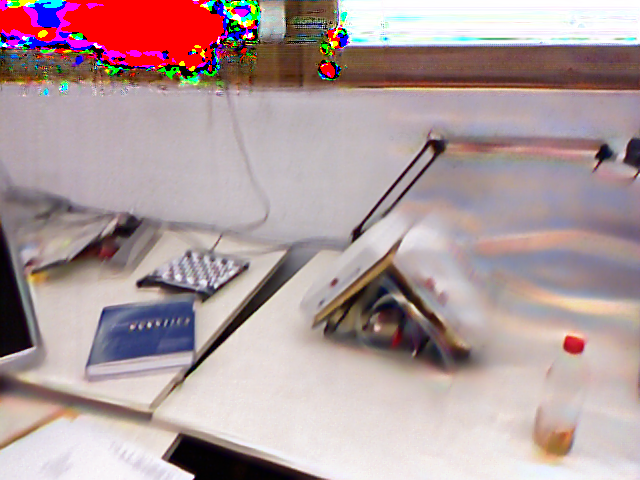} &
        \includegraphics[width=\sz\textwidth]{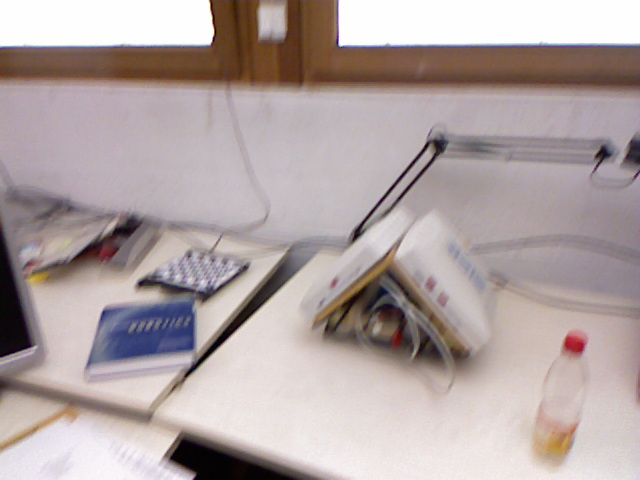} \\
        & Input & EVSSM~\cite{kong2025efficientvisualstatespace} & Ours\\
    \end{tabular}
    
    \caption{Comparison of input images and deblurred results}
    \label{fig:deblur_comparison}
\end{figure}
Table~\ref{tab:ate_mean} shows that insufficient training data prevents the network from learning rigorous intermediate frame extraction, causing the deblurring neural network to acquire 3D-inconsistent sharpening abilities. In complex scenes, Our* augmented solely with one synthetic blur sequence cannot extract reasonable intermediate frames or less-blurred intermediate frames in more general scenarios, thus resulting in potential 3D inconsistencies.


\subsection{Comparison to EVSSM}

We show additional results in comparison to the recent offline method EVSSM~\cite{kong2025efficientvisualstatespace}.
We used the publicly available weights of EVSSM which was trained on the GoPro dataset~\cite{Nah_2017_CVPR}. The results are shown in Figure~\ref{fig:deblur_comparison}. EVSSM struggles with motion blur from moving objects along their edges, producing strange color patches and noise as camera motion and object motion occur jointly. 
In contrast, \methodname{} decouples camera motion and object motion, and their separate processing makes our approach more robust.


\section{Design Details}
\subsection{Tracking Module:}
Tracking Module Algorithm~\ref{alg:blur_tracking} shows the structure of our tracking module. 
Here, $\Phi_{\text{mono}}(\cdot)$ denotes the monocular depth estimator~\cite{eftekhar2021omnidata} and $b_{\text{fail}} \in \{0, 1\}$ represents the failed flag of the deblur neural network (Sec.~\ref{sec:physics_deblur}).

\begin{algorithm}[b]
\renewcommand{\baselinestretch}{0.8}\selectfont
\caption{Robust Blur-Aware Tracking Pipeline}
\label{alg:blur_tracking}
\begin{algorithmic}[1]
\Require 
    Current\! RGB\! image $\text{I}_\text{t}$, 
    \! Pose history $\text{H} = \{ \text{T}_\text{t-1}, \text{T}_\text{t-2}, \dots \}$, 
    Deblurring failure flag $b_{\text{fail}}$
\Ensure 
    Current camera pose $\text{T}_\text{t}$, 
    Dense depth map $\text{D}_\text{t}$
\State \textit{1. Initial Depth Estimation}
\State $\text{D}_{\text{mono}} \leftarrow \Phi_{\text{mono}}(\text{I}_\text{t})$
\State \textit{2. Blur Evaluation and State Estimation}
\If {$b_{\text{fail}}$}
    \State $\text{V}_{\text{rel}} \leftarrow \text{T}_{t-1} \cdot \text{T}_{t-2}^{-1}$
    \State $\text{T}_\text{t} \leftarrow \text{T}_\text{t-1} \cdot \text{V}_{\text{rel}}$
    \State $\text{D}_\text{t} \leftarrow \text{D}_{\text{mono}}$
\Else
    \State $(\text{T}_\text{t}, \text{D}_{\text{ref}}) \leftarrow \text{Droid-SLAM}(\text{I}_\text{t}, \text{D}_{\text{mono}}, \text{T}_\text{t-1})$
    \State $\text{D}_\text{t} \leftarrow \text{D}_{\text{ref}}$
\EndIf
\State \Return $\text{T}_t, \text{D}_t$
\end{algorithmic}
\end{algorithm}

\subsection{Thresholds:}
Thresholds are an effective solution for handling various blur levels and are fixed for all experiments (Sec.~\ref{sec:experiment}), based on the statistical distribution of values in the large-scale dataset (Sec.~\ref{sec:supp_blur_quant}). During mapping, we employ a maximum iteration limit and robustly handle 7-12 consecutive blurry frames on the Deblur-NeRF dataset\cite{ma2022deblur}. The deblurring network (Sec.~\ref{sec:physics_deblur}) could identify the blurred frame implicitly with more computation time.

\begin{table}[h]
\setlength{\columnsep}{5pt}
    \setlength{\tabcolsep}{1pt}
    \renewcommand{\arraystretch}{0.6} 
    \centering
    \label{tab:ablation_threshold}
    \begin{tabular}{lcc}
        \toprule
        Method & PSNR [dB] & FPS \\
        \midrule
        \text{Ours\_no\_threshold} & 27.18 & 0.41 \\
        \textbf{Ours} & \textbf{29.46} & \textbf{0.74} \\
        \bottomrule
    \end{tabular}
\end{table}
%
\noindent
The table compares the average metrics to a variant without threshold on the fr1\_desk, fr2\_xyz and fr3\_office sequences, which relies on the deblurring network for judgment. 

\subsection{Notion Table:}
In brief, $\lambda$ is the weight for the loss terms of a single frame, excluding $\lambda_{\text{reg}}$ (Eq.~\eqref{eq:global_bundle_adjustment}). 
Here is a notion overview:
\begin{table}[h]
    \centering
    \renewcommand{\arraystretch}{1.0} 
    \setlength{\tabcolsep}{1pt}
    \scriptsize
    \resizebox{\linewidth}{!}{
    \begin{tabular}{@{}l l l@{}}
    \toprule
    \text{Symbol} & \text{Definition} & \text{Context} \\
    \midrule
    $\text{x}$ & Pixel coordinates $\text{x}=\text{(u, v)}$ on image plane & All \\
    $\text{f}$ & Focal length used in defocus kernel calculation & Eq.~\eqref{eq:blur_formation} \\
    $\text{d(t)}$ & Camera depth at time $t$ during exposure & Eq.~\eqref{eq:blur_formation} \\
    $\sigma_{\Delta}$ & Standard Deviation of the Score Differences & Sec.~\ref{sec:supp_blur_quant}\\
    $\text{C}, \text{D}_\text{r}$ & Rendered 2D RGB and depth image & Eq.~\eqref{eq:projected_gaussian} \\
    $\text{d}, \text{d'}$ & Original Gaussian depth($\text{d}$) and updated tracker's depth($\text{d'}$) & Eq.~\eqref{eq:update_gaussian} \\
    $\text{R}_\text{k}, \text{t}_\text{k}$ & Rotation and translation of the $\text{k}$-th virtual sub-frame & Sec.~\ref{sec:complex_blur} \\
    $\theta_\text{k}, \rho_\text{k}$ & Lie algebra parameters for trajectory interpolation & Sec.~\ref{sec:complex_blur} \\
    $\text{w}_\text{sharp/deblur/fail/f}$ & Weights for sharp, deblurred, and failed frames & Eq.~\eqref{eq:bundle_adjustment}-\eqref{eq:global_bundle_adjustment} \\
    $\text{F}, \text{L}_{\text{f}}$ & Union and Loss term of sharp, deblurred, and failed frames & Eq.~\eqref{eq:bundle_adjustment}-\eqref{eq:global_bundle_adjustment} \\
    $\text{a}_{\text{exposure}}, \text{b}_{\text{exposure}}$ & Learnable affine Scale and Shift parameters for exposure & Eq.~\eqref{eq:scale_loss} \\
    \bottomrule
    \end{tabular}
    }
    \label{tab:notation_clarification}
\end{table}

\subsection{Blur Quantification and Recognition Details}
\label{sec:supp_blur_quant}

This section provides further details on the dataset construction and metric evaluation introduced in Sec.~\ref{sec:blur_quant} of the main paper.

\boldparagraph{Dataset construction.}
\label{sec:dataset construction}
Our comprehensive blur detection benchmark incorporates real-world datasets alongside two custom semi-synthetic datasets. Following ``Image as an IMU''~\cite{chen2025imageimu}, we first apply the Eden~\cite{zhang2025eden} video interpolation model to the ScanNet~\cite{dai2017scannet} dataset to generate intermediate frames. To physically model motion blur, we convert images from sRGB to linear space using inverse gamma correction, synthesize motion blur by averaging $\text{N}$ interpolated frames, and then convert back to sRGB space:
\begin{equation}
\label{eq:reverse_rgb}
\text{I}_{\text{blur}} = \gamma^{-1}\left(\frac{1}{\text{N}} \sum_{\text{i}=\text{1}}^{\text{N}} \gamma(\text{I}_{\text{linear}}^{\text{(i)}})\right)
\end{equation} 
where $\gamma(\cdot)$ and $\gamma^{\text{-1}}(\cdot)$ represent the gamma correction and its inverse. The middle interpolated frame serves as the ground truth sharp frame to ensure physical correctness.

The second semi-synthetic dataset combines sequences from the RED~\cite{Nah_2019_CVPR_Workshops_REDS}, GoPro~\cite{Nah_2017_CVPR}, and ReplicaBlurry~\cite{girlanda2025deblur} datasets, with manual annotation to remove frames containing moving objects or inherent blur.

\boldparagraph{Blurry frames detection.}
To evaluate the 39 blur detection metrics, we designed a proxy measurement on a selected dataset with paired sharp and blurry images. The improvement score $\overline{\Delta}$ quantifies the magnitude of the quality difference between sharp and blurred images:
\begin{equation}
\overline{\Delta} = \frac{1}{\text{M}} \sum_{\text{i}=\text{1}}^{\text{M}} |\text{s}_{\text{sharp}}^{\text{(i)}} - \text{q}_{\text{blur}}^{\text{(i)}}|
\end{equation}
where $\text{q}$ denotes quality scores, and $\text{M}$ is the total number of sharp/blurry pairs.

Effect size uses Cohen's d~\cite{cohen1988statistical} to measure standardized separation: $\text{d} = \overline{\Delta} / \sigma_{\Delta} $.
Additionally, accuracy $a$ measures the percentage of correctly ranked blurry images.
The final consistency score combines accuracy and effect size by measuring $(\text{a} \times |\text{d}|) / \text{100}$.
We select the blur detection metric that achieves the highest consistency score as our blur detector, which is ARNIQA~\cite{agnolucci2024arniqa} (Table~\ref{tab:iqa_metrics}).

\begin{table}[h]
\small
\centering
\setlength{\tabcolsep}{6.8pt}
\begin{tabular}{lccc}
\toprule
      &            & Effect & Consistency\\
Metric & Acc. [\%] & size   & score \\
\midrule
ARNIQA-CSIQ~\cite{agnolucci2024arniqa} & 92.93 & \textbf{37.01} & \textbf{34.39} \\
MANIQA-KADID~\cite{yang2022maniqa} & \text{96.55} & 6.35 & 6.13 \\
CLIPIQA+~\cite{wang2023exploring} & 96.58 & 2.26 & 2.18 \\
MUSIQ~\cite{ke2021musiq} & 95.13 & 2.14 & 2.03 \\
BRISQUE~\cite{6272356} & 79.78 & 1.33 & 1.06 \\
\bottomrule
\end{tabular}
\caption{\text{Blur detection metric.} We compared the accuracy and consistency scores of 39 blur metrics and selected 5 representative ones for reporting. We chose the metric with the highest consistency score among them as our blur detector (Sec.~\ref{sec:blur_quant}).}
\label{tab:iqa_metrics}
\end{table}
\section{Additional qualitative results}
\begin{figure}[t]
    \centering
    \begin{tabular}{cc}        
        \includegraphics[width=0.45\columnwidth]{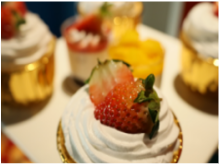} & 
        \includegraphics[width=0.45\columnwidth]{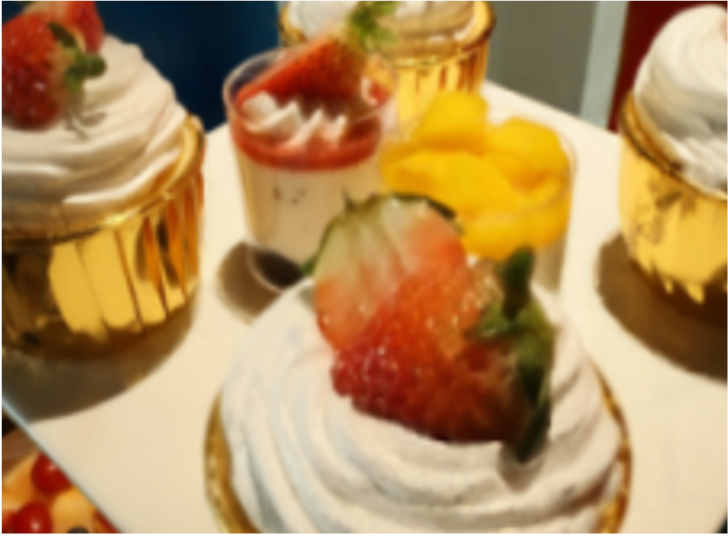} \\
        \includegraphics[width=0.45\columnwidth]{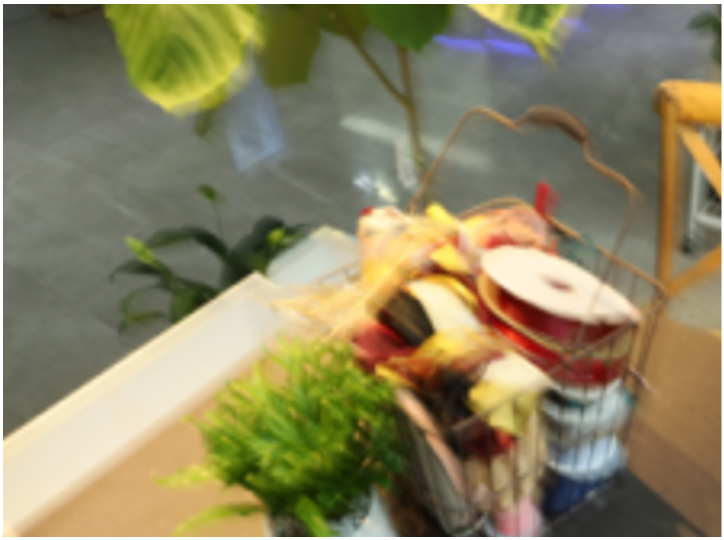} & 
        \includegraphics[width=0.45\columnwidth]{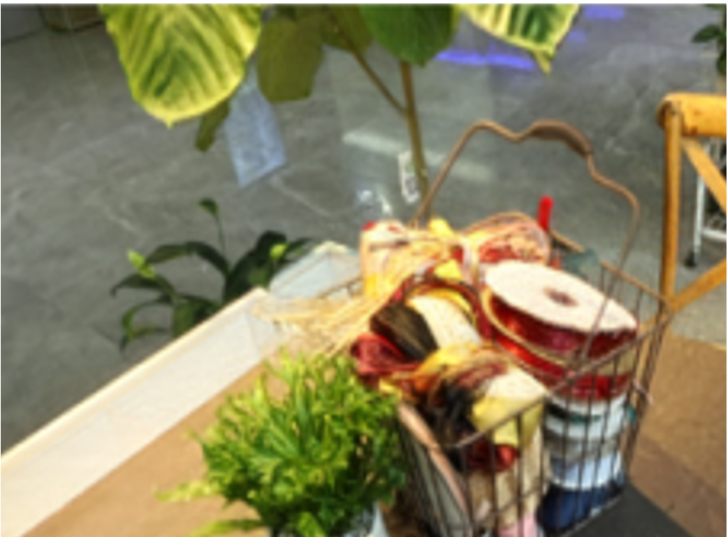} \\
        \includegraphics[width=0.45\columnwidth]{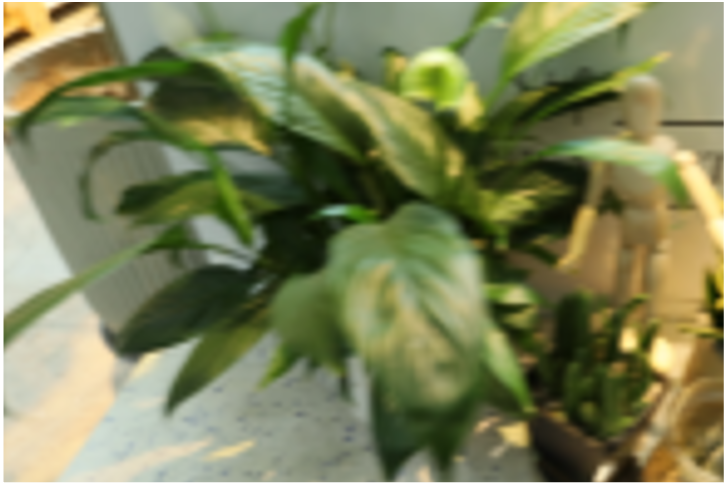} & 
        \includegraphics[width=0.45\columnwidth]{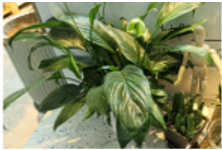} \\
        \includegraphics[width=0.45\columnwidth]{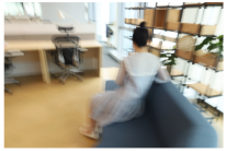} & 
        \includegraphics[width=0.45\columnwidth]{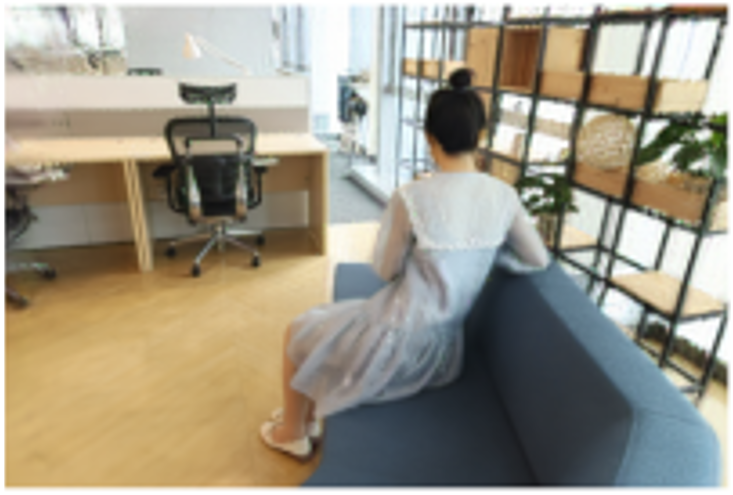} \\
        \includegraphics[width=0.45\columnwidth]{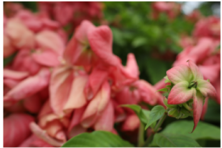} & 
        \includegraphics[width=0.45\columnwidth]{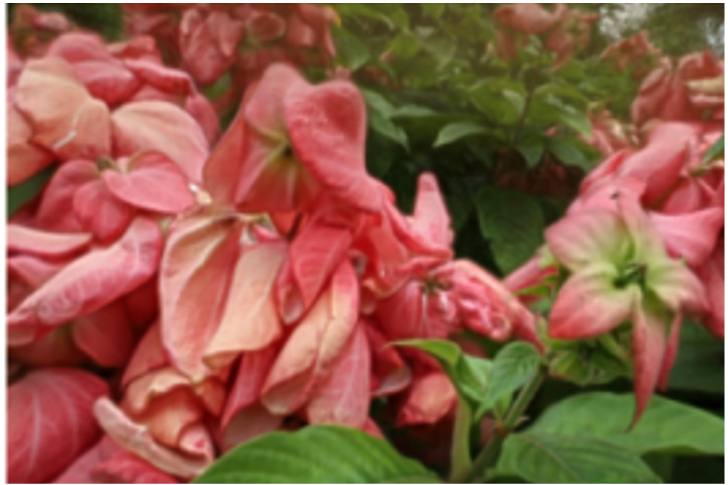} \\
        Input & Ours \\
    \end{tabular}
    \caption{\textbf{Qualitative results on various datasets.} The left column shows the input images, while the right column displays the rendered images of our reconstructed 3D models.}
    \label{fig:add_quli}
\end{figure}

We conducted additional qualitative experiments on various datasets to demonstrate the robustness of our method. As shown in Figure~\ref{fig:add_quli}, our approach exhibits improved detail recovery in various scenarios.
\section{Limitations and Future Work}

The physically constrained deblurring neural network training method proposed in Sec.~\ref{sec:physics_deblur} is unsuitable for images where the linear RGB color space cannot be accurately restored. These images typically undergo complex optimization algorithms by mobile phone pipelines, rendering them nearly impossible to reverse-engineer. Furthermore, while the proposed fallback mechanism for severe blur modeling renders multiple virtual sub-frames, the memory footprint grows explosively as the number of sub-frames increases. This memory constraint inherently limits its further expansion and scalability within online methodologies.

Nevertheless, this work proposes a reasonable engineering pipeline to implement a theoretical framework aimed at promoting progress within the community. Given the rapid development in the SLAM field, pure neural network-based approaches represent a highly promising direction for the further development of this theory.

